\documentclass[11pt, twoside]{article}
\usepackage{arxiv}

\usepackage{amsfonts}
\usepackage{amsmath}
\usepackage{comment}
\usepackage{amssymb}
\usepackage{amsthm}
\usepackage{bbm}
\usepackage{tabularx}
\usepackage{placeins}
\usepackage[caption=false]{subfig}
\usepackage{color,colortbl}
\usepackage[dvipsnames]{xcolor}
\usepackage{afterpage}
\usepackage{pdflscape}
\usepackage{hyperref}
\usepackage{cleveref}
\usepackage{acronym}
\usepackage[numbers]{natbib}
\usepackage{graphicx}
\usepackage{pifont}
\usepackage[font=footnotesize]{caption}
\usepackage{fancyhdr}
\usepackage{graphicx} % Required for inserting images
\usepackage{todonotes}
\usepackage{array}
\usepackage{url}
\usepackage{multirow}
\usepackage{float}
\usepackage{multicol}
\usepackage{mathtools}
\usepackage{schemata}
\usepackage{dsfont}
\usepackage{float}
\usepackage{algorithm}
\usepackage{algorithmicx}
\usepackage[noend]{algpseudocode}
\usepackage[inline]{enumitem}
\usepackage{booktabs}
\usepackage{amssymb}
\usepackage{listings}
\usepackage{xcolor}
\usepackage{comment}
\usepackage[toc,page]{appendix}
\usepackage{ifthen}
\usepackage{arydshln}
\usepackage{natbib}

\crefformat{footnote}{#2\footnotemark[#1]#3}

\definecolor{LightGray}{gray}{0.9}

\newcommand{\interior}[1]{%
  {\kern0pt#1}^{\mathrm{o}}%
}

\newcommand{\matrixnorm}[1]{{\left\vert\kern-0.25ex\left\vert\kern-0.25ex\left\vert #1 
    \right\vert\kern-0.25ex\right\vert\kern-0.25ex\right\vert}}

\theoremstyle{definition}

\theoremstyle{remark}

\theoremstyle{plain}

\title{An Interpretable Client Decision Tree Aggregation process for Federated Learning}

\author{Alberto Argente-Garrido \And 
        Cristina Zuheros \And 
        M. Victoria Luzón \And 
        \bigskip
        Francisco Herrera \And \\
A. Argente-Garrido, C. Zuheros and F. Herrera are with the \\Department of Computer Sciences and Artificial Intelligence, and the \\Andalusian Research Institute in Data Science and Computational Intelligence (DaSCI)\\ 
 University of Granada, 18071 Granada, Spain.  \\
        \bigskip  \textit{aargente@ugr.es; czuheros@ugr.es; herrera@decsai.ugr.es}\\
M.V. Luzón is with the Department of Software Engineering, and the\\Andalusian Research Institute in Data Science and Computational Intelligence (DaSCI)\\ University of Granada, 18071, Granada, Spain. \\\textit{luzon@ugr.es}}

\hypersetup{
pdftitle={An Interpretable  Client Decision Tree Aggregation process for Federated Learning},
pdfsubject={Federated Learning},
pdfauthor={Alberto Argente-Garrido, Cristina Zuheros, M.V. Luzón, Francisco Herrera},
pdfkeywords={Federated Learning, Decision Trees, Interpretability, Aggregation Process, Data Privacy},
}

\date{\today}

\renewcommand{\headeright}{}
\renewcommand{\undertitle}{}
\algdef{SE}[SUBALG]{Indent}{EndIndent}{}{\algorithmicend\ }%
\algtext*{Indent}
\algtext*{EndIndent}

\begin{document}

\maketitle

%\begin{frontmatter}

%\author[dasciaddress]{Juan Luis Su\'arez\corref{corresp}}
%\ead{jlsuarezdiaz@ugr.es}
%\author[dasciaddress]{Salvador Garc\'ia}
%\ead{salvagl@decsai.ugr.es}
%\author[dasciaddress]{Francisco Herrera}
%\ead{herrera@decsai.ugr.es}

%\cortext[corresp]{Corresponding author}
                  
%\address[decsaiaddress]{Department of Computer Sciences and Artificial Intelligence, University of Granada, C/ Periodista Daniel Saucedo Aranda, s/n, 18014, Granada, Spain}
%\address[dasciaddress]{DaSCI, Andalusian Research Institute in Data Science and Computational Intelligence, University of Granada, Granada, Spain}

\begin{abstract}

Trustworthy Artificial Intelligence solutions are essential in today's data-driven applications, prioritizing principles such as robustness, safety, transparency, explainability, and privacy among others. This has led to the emergence of Federated Learning as a solution for privacy and distributed machine learning. While decision trees, as self-explanatory models, are ideal for collaborative model training across multiple devices in resource-constrained environments such as federated learning environments for injecting interpretability in these models.
Decision tree structure makes the aggregation in a federated learning environment not trivial. They require techniques that can merge their decision paths without introducing bias or overfitting while keeping the aggregated decision trees robust and generalizable.
In this paper, we propose 
an Interpretable Client Decision Tree Aggregation process for Federated Learning scenarios that keeps the interpretability and the precision of the base decision trees used for the aggregation. 
This model is based on aggregating multiple decision paths of the decision trees and can be used on different decision tree types, such as ID3 and CART.
We carry out the experiments within four datasets, and the analysis shows that the tree built with the model improves the local models, and outperforms the state-of-the-art.

\end{abstract}

%\begin{keyword}
%    Distance Metric Learning \sep Classification \sep Mahalanobis Distance \sep Dimensionality \sep Similarity
%\end{keyword}

%\end{frontmatter}

%\pagebreak
%\tableofcontents
%\pagebreak

% \linenumbers

%% SECTION 1 - INTRODUCTION
\section{Introduction}\label{sec:introduction}

Today's data-driven applications require the search for Trustworthy Artificial Intelligence (TAI) \cite{TAIScottThiebes2020} solutions that not only focus on the accuracy of the model, leading traditional Machine Learning (ML) to approach Deep Learning (DL) solutions that can capture the complexity of the data. These approaches have led to a disregard for data privacy and data protection. To solve these problems, TAI requires to create a responsible AI system \cite{DIAZRODRIGUEZ2023101896} that goes beyond accuracy and focuses on the principles of AI trustworthiness, including robustness, safety, transparency, and explainability, among others \cite{trustworhtyPrinciples, weipingExplain}. To address this challenge, Federated Learning (FL) emerges as a solution at the intersection of data privacy and distributed ML \cite{ALHUTHAIFI2023833}, particularly relevant in the context of privacy and data protection, whether trustworthiness is required \cite{RiskFree2023}. Transparency and explainability are crucial in a trustworthy system, which leads to the use of self-explanatory models.

Decision Trees (DTs) are an example of self-explanatory models because their structure is inherently interpretable \cite{ROKACH2016111}, facilitating transparency and trust among stakeholders in FL environments. This interpretability is crucial, particularly in sensitive fields like healthcare or finances, where understanding the behavior behind the model is essential for stakeholder acceptance. DTs are lightweight and computationally efficient, which makes them well-suited for deployment on resource-constrained devices, often found in FL. The simplicity and low computational requirements of DTs enable efficient model training and inference without compromising device performance or requiring extensive computing resources. These properties make DT a valuable asset in an FL environment. We can find in the literature approaches for DT \cite{truex2019hybrid} and for decision forests \cite{federatedforest_vfl_2020, biorf2022, BOFRF2022}.

Aggregating DTs in an FL environment presents particular challenges due to the structure and characteristics of DTs, along with the intrinsic nature of FL. %Other potential models, such as linear models or neural networks, which are parameter-based models, are updated via gradient-based methods \cite{LIU202014, NI2024119784, ZHAO2024119873, NANOR2023119725}. 
In the case of DTs, they consist of hierarchical structures with branching decision paths, making the direct aggregation not trivial. Contrary to parameter-based models, aggregating DTs requires techniques that can merge decision paths across the multiple nodes of the FL environment while preserving the interpretability and integrity of the resulting global model. DTs tend to be sensitive to data distribution and variance, which can vary a lot across the distributed nodes in FL scenarios. Aggregating DTs from diverse datasets without introducing bias or overfitting is also a considerable challenge. Methods for aggregating DTs need to ensure that the resulting global model is robust and generalizable across all nodes. Addressing these challenges requires the development of a specialized aggregation process tailored to the complexities of the DT models and the FL environments.

To address this challenge, we propose a novel aggregation process for DT in FL environments namely Interpretable Client Decision Tree Aggregator For FL, called ICDTA4FL process. This process filters the low-quality DTs and aggregates the decision paths of these DTs built at different nodes into a global DT, keeping the structure, interpretability, and privacy of the data while improving the performance of the local DTs. Rather than training for multiple rounds as with parametric-based methods, a single round is required in which DTs are decomposed into branches that are merged to form the global DT. The ICDTA4FL process works for multiple DT types. We present:
\begin{enumerate*}[label=(\arabic*)]
    \item the ICDTA4FL-ID3 model which uses ID3 \cite{Quinlan1986} as the base model for building the global DT, and
    \item the ICDTA4FL-CART model that uses CART \cite{breiman2017classification} as the base model for building the global DT.
\end{enumerate*}

We conduct several experiments to ensure the performance of the ICDTA-4FL process. We compare our results against those from the state-of-the-art models.
We show the capability of the ICDTA4FL process to adapt to multiple DT algorithms, using ID3 and CART, with four different tabular datasets. We show how the global model obtained with the ICDTA4FL process improves the local DT model from the nodes, being scalable and generalizable. We analyze the results of the global model, and how it improves the local model while keeping the inherent interpretability and complexity from the original DTs. The ICDTA4FL process filters client's DTs to delete possible noise added to the model by clients with low-performance DTs. The selected filter is important because it helps keep the model performance obtained when the number of clients increases.

The rest of this paper is organized as follows: Section \ref{sec:background} describes the DT and FL paradigm. Section \ref{sec:proposal} describes the proposed ICDTA4FL process and models in detail. Section \ref{sec:expframe} details the experimental framework carried out to prove the performance of the ICDTA4FL process. Section \ref{sec:exresanlysis} analyzes the results obtained in the experimental case study by the ICDTA4FL process. Finally, Section \ref{sec:conclusions} presents some concluding remarks and future work.

%% SECTION 2 - BACKGROUND
\section{Background}\label{sec:background}

This section presents the basic concepts related to the ICDT4FL process. Section \ref{subsec:bgfl} focuses on the concepts of FL. Section \ref{subsec:bgdt} introduces the work concerning DTs and % Section \ref{subsec:bgexpldt} 
emphasizes the interpretability of DTs. Finally, Section \ref{subsec:bgdtfl} is an introduction to the work on both DTs and FL.

\subsection{Federated Learning}\label{subsec:bgfl}

FL \cite{konevcny2016federated, yang2019federated, silva2023towards, tutorialFL2024} is a distributed learning paradigm designed to facilitate model learning from decentralized data without the need to centralize data on a server. This approach ensures that local data remains exclusively on the device where it was originally collected, thereby guaranteeing data privacy. FL has gained significant attention for its effectiveness in addressing privacy concerns and improving the efficiency of distributed learning \cite{chen2023building, zong2023fedcs, ALHUTHAIFI2023833}.\\

FL accommodates numerous participants, each with their own data sources, making it particularly well-suited to scenarios where security and data privacy are mandatory, and there is also a need to improve performance and accuracy \cite{ALHUTHAIFI2023833}.This has positioned FL as an important area in AI, attracting the interest of researchers, developers, and data scientists in the ML community to explore both theoretical and practical applications involving sensitive data. \\

% \todo[inline]{Añadir HFL y VFL}

Horizontal Federated Learning (HFL) and Vertical Federated Learning (VFL) represent two different approaches within the FL paradigm. In HFL, multiple clients share the features of the problem but different data samples. On the other hand, VFL involves clients that share data samples but have different sets of features. Both approaches offer unique solutions to privacy-preserving collaborative learning across distributed data sources, satisfying diverse scenarios and data structures in real-world applications.

\subsection{Decision Trees and Interpretability}\label{subsec:bgdt}

% \todo[inline]{Alberto: Dos paragraphs para diferenciar de árboles de decisión individuales como CART/ID3/C4.5 de modelos de ensembles como Random Fores/Gradient Boosting Decision Trees}

A DT is a predictive model that maps different decisions in a branching structure. DTs are very powerful approaches used in ML to solve diverse problems, such as classification or regression problems. DTs are very popular in ML because of their simple structure, which makes the model explainable and understandable, but they still suffer from several drawbacks, such as instability. Two well-known DT algorithms are: \begin{enumerate*}
    \item Classification and Regression Trees (CART) \cite{breiman2017classification} is a versatile algorithm suitable for both classification and regression tasks. It constructs binary trees by recursively partitioning the dataset based on feature conditions. 
    \item Iterative Dichotomiser 3 (ID3) \cite{Quinlan1986} is primarily designed for classification problems. It employs a top-down, recursive approach, selecting the best attribute at each node based on information gain.
\end{enumerate*}\\

% Two well-known DT algorithms are Classification and Regression Trees (CART) \cite{breiman2017classification} and Iterative Dichotomiser 3 (ID3) \cite{Quinlan1986}. CART is a versatile algorithm suitable for both classification and regression tasks. It constructs binary trees by recursively partitioning the dataset based on feature conditions. On the other hand, ID3 is primarily designed for classification problems. It employs a top-down, recursive approach, selecting the best attribute at each node based on information gain.\\

To mitigate the inherent limitations of individual DTs, ensemble learning techniques are used to improve the performance. %, and combining multiple models for improved performance, have gained prominence. 
Notable ensemble models include Random Forest (RF) \cite{breiman2001random} and Gradient Boosting Decision Trees (GBDT) such as LightGBM \cite{NIPS2017_6449f44a} and XGBoost \cite{Chen_2016, GONZALEZ2020205}. RF is a bagging technique for DTs that enhances  
%, an ensemble of decision trees, introduces randomness during tree-building, enhancing 
generalization and %reducing 
reduces overfitting. LightGBM and XGBoost, are extreme gradient boosting algorithms, that employ a sequential boosting technique, refining the model's performance through iterative optimization. However, ensemble models lose the explainability from the inherited structure of the individual DTs. This makes them less desirable when interpretability is a key factor in solving a problem. \\

% Ensemble models capitalize on the strengths of individual decision trees while mitigating their weaknesses. By aggregating predictions from multiple trees, these models often yield enhanced accuracy, robustness, and generalization to diverse datasets. Embracing both the simplicity of decision trees and the power of ensemble methods.\\

% \subsection{Interpretability of Decision Tree-based models}\label{subsec:bgexpldt}

% \todo[inline]{Alberto: Hablar de que los modelos de árboles de decisión son interpretables por su estructura, pero que los modelos de ensembles, sobre todo aquellos que usan muchos árboles, pieden las interpretabilidad (así se justifica luego que se agrupen todos los árboles en 1).}

The interpretability and explainability of ML models are key considerations in the context of ML and FL \cite{barredo2020}. DTs have innate interpretability due to their hierarchical and rule-based structure \cite{ROKACH2016111}. This structure provides a clear and intuitive representation of the decision process, that is accessible to both experts and non-experts. This becomes crucial in an FL environment, where models are trained on decentralized devices \cite{rodriguez2020federated}. \\

The structure of DTs makes them preferable to parameter-based models like neural networks, which can hide relationships between input features and predictions. %\cite{lecun2015deep}.
Neural networks excel at finding patterns but lack transparency, raising concerns in applications where interpretability is crucial \cite{barredo2020}. A model must therefore meet the seven requirements of trustworthiness to be used \cite{DIAZRODRIGUEZ2023101896}, making DT a suitable option.
% Also, due to the new EU regulations \cite{Act_2021}, where a model has to be explainable to be used, DT might be a suitable option. DTs can help experts who need clear insights to take the model's prediction into account.

% This interpretability and explainability inherited by the tree's structure make this method be selected above black box models, such as neural networks, as the relationships between input features and output predictions are often obscured by the complex, non-linear transformations occurring within the network \cite{lecun2015deep}. While neural networks excel in capturing patterns and representations, their lack of transparency raises concerns, particularly in applications demanding accountability and interpretability \cite{barredo2020}. In federated learning scenarios, the opacity of neural networks may not be desired in those fields in which the interpretability of the results is required in the decision-making process, such as healthcare, where a decision tree might be a suitable option for the experts to use a model that can help the expert to take the final decision. \\

\subsection{Decision Trees in Federated Learning}\label{subsec:bgdtfl}

% Researchers are actively exploring techniques to enhance the interpretability of decision tree-based models in FL, capitalizing on their transparency without sacrificing performance. FL proves versatile in training both single and multiple decision trees, including popular variants like GBDT and RF. In practice, both RF and GBDT have been used in Horizontal-FL (HFL) and Vertical-FL (VFL). \\

DTs have been successfully applied in the FL scenario, proving the versatility of FL in training both single and multiple DTs. In practice, ensemble methods such as RF \cite{biorf2022, federatedforest_vfl_2020, revrfieee} and GBDT \cite{eflBoost2022, opboost2022acm} have been used in both HFL and VFL over single tree methods like ID3 \cite{truex2019hybrid}. \\

% Paragraph of Bagging in both HFL and VFL
In the state-of-the-art we find multiple bagging approaches such as Hauschild et al. \cite{biorf2022}, that build an RF in an HFL environment for medical data, and each node builds an RF that contains k-decision trees. Then k-decision trees are randomly sampled from among those available. Yang Liu et al. \cite{federatedforest_vfl_2020} built each tree from the RF collaboratively between the participating nodes. On the other hand, Yang Liu et al. \cite{revrfieee} build an RF collaboratively. However, if a client drops out of the training process, all data associated with that client is removed from the model and any tree the client contributed to is rebuilt. These approaches involve a limited number of participant nodes when training the global model. \\

% \todo[inline]{Alberto: Cambiar las citas, que no sea In [8], o Los autores de [8], o [8], sino que sea Autor et Al. [8]}

% Paragraph of Boosting in both HFL and VFL
On the other hand, boosting approaches like Kewei Chen et al. \cite{secureboost2021} propose a VFL framework for boosting, where it aligns the data under privacy constraints, and then trains a GBDT while keeping all the training data secured from other multiple private parties. Thus, there is a high communication cost between nodes, and other approaches work towards reducing it. Weijing Chen et al. \cite{chen2021secureboost} improve SecureBoost by incorporating various optimizations for ciphertext calculations and engineering optimizations. Xiaochen Li et al. \cite{opboost2022acm} present another VFL framework for boosting, OpBoost, that looks forward to optimizing the communication and computation cost of other methods, by optimizing the solution based on distance-based Local Differential Privacy (dLPD).

Yamamoto et al. \cite{eflBoost2022} propose an HFL framework for boosting that optimizes global and local computations to reduce communication costs. Gencturk et al. \cite{BOFRF2022} propose a boosting framework in HFL creating an RF node, and then it uses all the trees from each RF to update the weights of the local instances from each node. Qinbin Li et al. \cite{li2020practical} focus on constructing a global hash table to align instances without sharing any private data, followed by training a GBDT where each tree is built by one node at a time. These approaches lack large-scale federated scenarios, similar to bagging approaches.\\

Truex et al. \cite{truex2019hybrid} proposes the Federated-ID3 model, this is, the ID3 \cite{Quinlan1986} adapted to a HFL environment. This model differs from the bagging and boosting strategies explained above. Section \ref{subsec:sota} describes this model in more detailincluidng in section 5 a comparative study with our proposal, including in Section \ref{sec:expframe} a comparative study with our proposal. \\

%% SECTION 3 - DISTANCE METRIC LEARNING ALGORITHMS
\section{An Interpretable  Client Decision Tree Aggregation process for Federated Learning and models}\label{sec:proposal}

% Interpretable Client FLDTAggModel (ICFLDTA model)

This section presents the Interpretable Client Decision Tree Aggregator For Federated Learning (ICDTA4FL).
% ICDTA4FL
It proposes an FL aggregation process for DT models that we particularize to ID3 and CART-based models. Through this section, we first present the general aggregation process, and then we specify the differences in aggregation between ID3 and CART. Thus, this section is composed of three subsections. Subsection \ref{subsec:aggmethod} explains the general aggregation process. Subsection \ref{subsec:id3agg} is about how to adapt the ICDTA4FL to the ID3 algorithm, getting the ICDTA4FL-ID3 model. Subsection \ref{subsec:cartagg} is about how to adapt the ICDTA4FL to the CART algorithm, getting the ICDTA4FL-CART model.

% an FL aggregation method for DT models. Section \ref{subsec:aggmethod} presents the aggregation steps followed to aggregate multiple decision trees. Section \ref{subsec:id3agg} exposes how the methodology adapts to the ID3 algorithm. Section \ref{subsec:cartagg} shows how to adapt the methodology to the CART algorithm.
% ID3 and CART-based models. Through this section, we first present the general aggregation methodology for both tree types, and then the differences in aggregation between ID3 and CART will be specified. This section is composed of three subsections. In the first subsection \ref{subsec:aggmethod} the general methodology followed in this aggregation process is explained. The second subsection \ref{subsec:id3agg} is about how to adapt the methodology to the ID3 algorithm. The last subsection \ref{subsec:cartagg} is about how to adapt the methodology to the CART algorithm.

\subsection{Interpretable Client Decision Tree Aggregation For Federated Learning process (ICDTA4FL process)}\label{subsec:aggmethod}

We propose an aggregation process for DTs in an FL environment, where nodes cannot share their data but still want to train an FL model% , which is crucial in fields such as healthcare \cite{ishealthcare2023}.\\
% A decision tree model can be decomposed into a set of decision rules, and this set of rules can be organized to build a decision tree\cite{SAGI2020124}. In Section \ref{sec:background}, we discuss federated learning decision tree models that are built through the collaboration of the nodes involved in the training, in both Horizontal Federated Learning (HFL) and Vertical Federated Learning (VFL). These approaches typically require a lot of client-server communication to build the final model, and they seek to create a single tree or a set of trees at once.\\
A DT model can be decomposed into a set of decision rules, and this set can be structured to build a DT \cite{SAGI2020124}. In Section \ref{sec:background}, we explore the Federated Learning Decision Tree models developed through collaborative efforts among the nodes involved in the training in HFL. These approaches often require extensive communication between clients and a server to build the final model, aiming to build either a singular tree or a set of trees.\\

We propose a decision-tree-independent aggregation model for HFL where there are few client-server communications. The proposed ICDTA4FL process relies on the ability to merge DTs \cite{SAGI2020124}. The Algorithm \ref{alg:alg1} shows the ICDTA4FL workflow.
It %only 
requires one round for training the global model, and it 
% only 
needs four client-server communication steps, lowering the communication costs of the training and making it resistant to possible connection errors during the training phase.\\

% The aggregation process works similarly to traditional FL aggregators, such as FedAvg\cite{fedavg}. The Algorithm \ref{alg:alg1} shows the aggregation workflow, which is decomposed into the tasks described below.\\

In the ICDTA4FL process, we consider a set of clients $C=\left\{C_{1},..., C_{n}\right\}$, which only train their local model and a server that acts as a communication moderator and as an aggregator, denoted by \textit{Server}. Each client $C_{i}, \forall i=1,\dots,n$, has its local data, $D_{i}, \forall i=1,\dots,n$, which is never shared with other clients or with the server. Subsequently we describe the ICDTA4FL workflow.\\

\begin{algorithm}
\caption{The ICDTA4FL process}\label{alg:alg1}

\begin{algorithmic}[1]
\State \textbf{\textit{Client's side}}
% \For{steps = $1$ to $N$}
    \Indent
    \For{ $C_{i}$; i=$1, \dots, n$}%$i$ to $N\_client$}
        \State $C_{i}$ $\leftarrow$ train a decision tree, $LocalDT_{i}$, with its local data $D_{i}$.
        \State Send $LocalDT_{i}$ to the Server.
    \EndFor
    \EndIndent
\State \textbf{\textit{Server's side}}
    \Indent
    \State Send the received trees to the clients.
    \EndIndent
\State \textbf{\textit{Client's side}}
    \Indent
    \For{$C_{i}$; i=$1$ to $n$}
        \State $C_{i}$ $\leftarrow$ evaluate the local DTs, $C_{k}LocalDT_{i}$, $k=1, \dots, n; i \neq k$ % evaluate all the trees on $D_{i}$.
        \State Send the evaluation metrics to the server.
    \EndFor
    \EndIndent
\State \textbf{\textit{Server's side}}
    \Indent
    \State Delete the trees that do not surpass a filter selected for the metrics.
    \State Extract the rules for selected decision trees.
    \State Aggregate the rules applying the Cartesian product.
    \State Build a global decision tree, $GlobalDT$ with the aggregated rules.
    \State Send $GlobalDT$ and the aggregated rules to the clients.
    \EndIndent
\State \textbf{\textit{Client's side}}
    \Indent
    \For{$C_{i}$; i=$1$ to $n$}
        \State Evaluate the $GlobalDT$ with its local data $D_{i}$
    \EndFor
    \EndIndent
\end{algorithmic}
\end{algorithm}

\paragraph{(\textbf{Client's side}) Build and send local decision tree}

There is a set of clients $C=\{C_1, \dots, C_n\}$ and a server denoted by $Server$. Each client $C_i$, $\forall i=1, \dots, n$ has its local data, $D_{i}$, that is never shared with other clients or with the server. The client $C_{i},\forall i = 1,\dots,n$ builds a local DT $LocalDT_{i}$ using the dataset $D_{i}$. After creating the DT, the client $C_{i}$ sends it to the server.

\paragraph{(\textbf{Server's side}) Send the local DTs to the clients}
%\paragraph{Send decision trees to the clients and evaluate them}

DTs tend to overfit the data they are trained on, thus it can affect the aggregation of the local trees, as they may add noise to the global model. To address this problem, the server resends all the DTs it receives to the clients and asks each client to evaluate the local DT from the rest of the clients on its data ($D_{i}$).

\paragraph{(\textbf{Client's side}) Evaluate the models from other clients}
The client $C_i,$ $i=1, \dots, n$ receives the local DT originally generated by client $C_k,$ $k=1, \dots, n$; $k\neq i$. We denote such a DT by $C_kLocalDT_i$. Each client $C_i$ evaluates the $n-1$ local DTs $C_kLocalDT_i$ on their private data $D_i$ and sends the achieved metrics to the server.

\paragraph{(\textbf{Server's side}) Build the global decision tree}
The Server has to build a global DT and send it to the clients. This process is depicted in the following tasks:

% \todo[inline]{Alberto: Terminar itemize}

\begin{itemize}
    \item \textbf{Delete lower performing trees:}
    % Once the server has all the evaluations, it can prune the trees to keep only those that surpass a predefined threshold. By default, the server uses the mean of the evaluations, keeping the trees whose evaluation is above the mean. The server can adjust the threshold to keep more or fewer trees, depending on its needs. This is an interesting approach, as it can be seen as a filter for clients with low-quality data, or for clients that may try to send a tree with low performance in order to affect the result of the final model.\\
    The Server has $n\times n$ evaluation metrics associated with the local DTs, which we denote by $EvalC_kLocalDT_i$, $i=1, \dots, n$; $k=1, \dots, n$. We compute the mean value of the metrics for each $LocalDT_{i}$, $i=1, \dots, n$ getting $EvalLocalDT_i$ = $\sum_{k=1}^{n}{EvalC_kLocalDT_i}$. We apply a filter to discard those trees whose $EvalLocalDT_{i}$ does not surpass the filter. We use the mean as the default filter, but others such as the percentile and the median can be used. Filtering helps reduce bad DTs that can add noise to the global DT. This step is essential when the number of clients increases, and when the data is highly distributed.
    % We discard those trees whose $EvalLocalDT_i$ does not reach the average value $\sum_{i=1}^{n}{EvalLocalDT_i}$. 
    \item \textbf{Get the rules from trees}: The server obtains the rules from each DT. Each rule is created from a leaf node to the root of the tree, so the rule contains all the associated constraints. Let $RS_{i}$ be the rule set associated with the DT $LocalDT_{i}$, $i=1, \dots, n$. This rule set consists of rules of the form $(r_{ij}, \hat{y}_{r_{ij}})$, where $r_{ij}$ is the condition of the rule and $\hat{y}$ is a vector containing the probability distribution of each of the $Y$ classes \cite{SAGI2020124}. $\hat{y}$ consists of a K-dimensional vector, where K is the number of classes available in the dataset, and each element of the vector corresponds to the probability of a particular class.
    \item \textbf{Aggregate the rules}: % Rules are aggregated by taking the Cartesian product between them, preceded by a compatibility check to ensure that they do not contradict each other. In our proposal, we demonstrate the aggregation of two types of trees, specifically ID3 and CART. The distinctions in these aggregation processes between the two tree types are detailed in the subsequent sections (\ref{subsec:id3agg} and \ref{subsec:cartagg}). Each rule is associated with a probability distribution of classes, and when two rules are combined, their distributions are summed.

    When aggregating two rules sets, $(r_{1j}, \hat{y}_{r_{1j}})$ and $(r_{2j}, \hat{y}_{r_{2j}})$, we employ the Cartesian product, resulting in the merging $(r_{1j}\wedge r_{2j}, \hat{y}_{r_{1j}} + \hat{y}_{r_{2j}})$. The sum of two probability vectors is done by mapping each element of the vector to its corresponding element. The compatibility of these rules is rigorously verified to ensure that they do not contradict each other, i.e., they are not incompatible. Depending on the DT type, this validation may vary 
    In the ICDTA4FL processS, we present %demonstrate
    the aggregation of two types of trees that serve as the basis for more complex models:
    \begin{enumerate*}[label=(\arabic*)] 
        \item ID3, since it is a classical DT learning algorithm that has been adapted to federated environments and 
        \item CART, since it is a cornerstone algorithm that serves as the basis for ensemble methods such as GBDT and RF.
    \end{enumerate*} In Section \ref{subsec:id3agg} we present how to validate the compatibility between two rules for the ICDTA4FL-ID3 model, and how the aggregation of those rules is done. We then do the same for the ICDTA4FL-CART in Section \ref{subsec:cartagg}.
    % We present how is conducted the aggregation of the rules for ID3 and CART in Sections \ref{subsec:id3agg} and \ref{subsec:cartagg}, respectively.

%     \todo[inline]{Alberto: Este itemize llevarlo a la subsección propia, así se puede especificar mejor en cada uno y se puede extender más la explicación.}
%    \begin{itemize}
%        \item \textbf{ID3 rules compatibility}: To decompose an ID3 tree into association rules, we identify the antecedent and consequent of each rule. The antecedent of a rule is the condition that must be met for the consequent to be true. The consequent of a rule is the prediction that is made if the antecedent is true. Therefore, to identify the antecedent and consequent of each rule, we simply need to traverse the tree from the root node to the leaf node, collecting the conditions at each node. To merge two association rules, A and B, we need to ensure that no antecedent of A is present in the antecedents of B. This is because if two rules have overlapping antecedents, then they are essentially saying the same thing.
%        \item \textbf{CART rules compatibility}: In the other hand, a CART is a binary tree the consequent of a rule doesn't have to be a specific value, but it is split as an interval, i.e., $age<=50, age>50$. To merge association rules, A and B, we need to ensure that if two antecedent-consequent pairs have the same consequent, then we can merge them if their antecedents are disjoint. For example, we cannot merge the rules $Age <= 18 => Youth$ and $Age > 18 => Youth$.
% \end{itemize}
    \item \textbf{Build the global DT using the aggregated rules}: The server aggregates a rule set containing rules and class probabilities to build a global DT $GlobalDT$. The tree's structure depends on the tree algorithm used by the clients. If the clients build a binary tree, the global tree will be binary. Otherwise, if the clients are using a different tree algorithm, then the global tree will have the same structure as the algorithm used by the clients.

    To initialize the global DT, the Server creates a root node containing all the rules from the rule set. To split a node, the Server selects a feature $f_{i} \in F$ that maximizes a specific metric, depending on the considered type of tree. The feature $f$ is calculated using the rules available in that node. The Server then splits the rule set into two subsets based on the value of the selected feature. Each child of the node will contain those rules that fulfill the constraints, and the process of building the global DT stops when the conditions are met. Specifically, the splitting process is finished when the class with the highest probability is consistent across all rules in the node,  or when the best splitting rule fails to reduce the number of rules in at least one child. This step is tree-dependent and is specified for the ICDTA4FL-ID3 model in Section \ref{subsec:id3agg}, and in Section \ref{subsec:cartagg} for the ICDTA4FL-CART model.

    %\item  \textbf{Evaluate the server on global data}: The $Server$ has global test data, inaccessible for the clients, and then probe the model with this data. Thus, we can test if the global model works not only with the local data from the clients but also with new data.
    \item \textbf{Sends the global DT to the clients}: The $Server$ sends the $GlobalDT$ and the aggregated rules to the clients.
    %\item \textbf{$Clients$ evaluate the global tree on their local data}: The client \czm{$C_i$, $i=1, \dots, n$} evaluates the global decision tree $GlobalDT$ on his/her own dataset $D_i$.
\end{itemize}

\paragraph{(\textbf{Client's side}) Evaluate the global decision tree} The client $C_i$, $i=1, \dots, n$ evaluates the global DT $GlobalDT$ on his/her own data $D_i$. Clients should consider this global DT to predict new data.

\subsection{ID3 rules aggregation process and tree building (ICDTA4FL-ID3 model)}\label{subsec:id3agg}

In the previous Section \ref{subsec:aggmethod} we present the ICDTA4FL process, which is capable of aggregating trees without the need to share any information between the clients, only the model. In this section, we present the specific steps needed for the ICDTA4FL process to work on an ID3 DT model, \textit{i.e.}, all the clients train locally in an ID3, namely the ICDTA4FL-ID3 model. There are two steps that are dependent on the tree: the aggregation of the rules and the construction of the global DT using these aggregated rules.

% \todo[inline]{Alberto: Escribir la sección usando paragraphs, una para la agregación de reglas y otra para la construcción del árbol.}

\paragraph{Aggregate the rules}

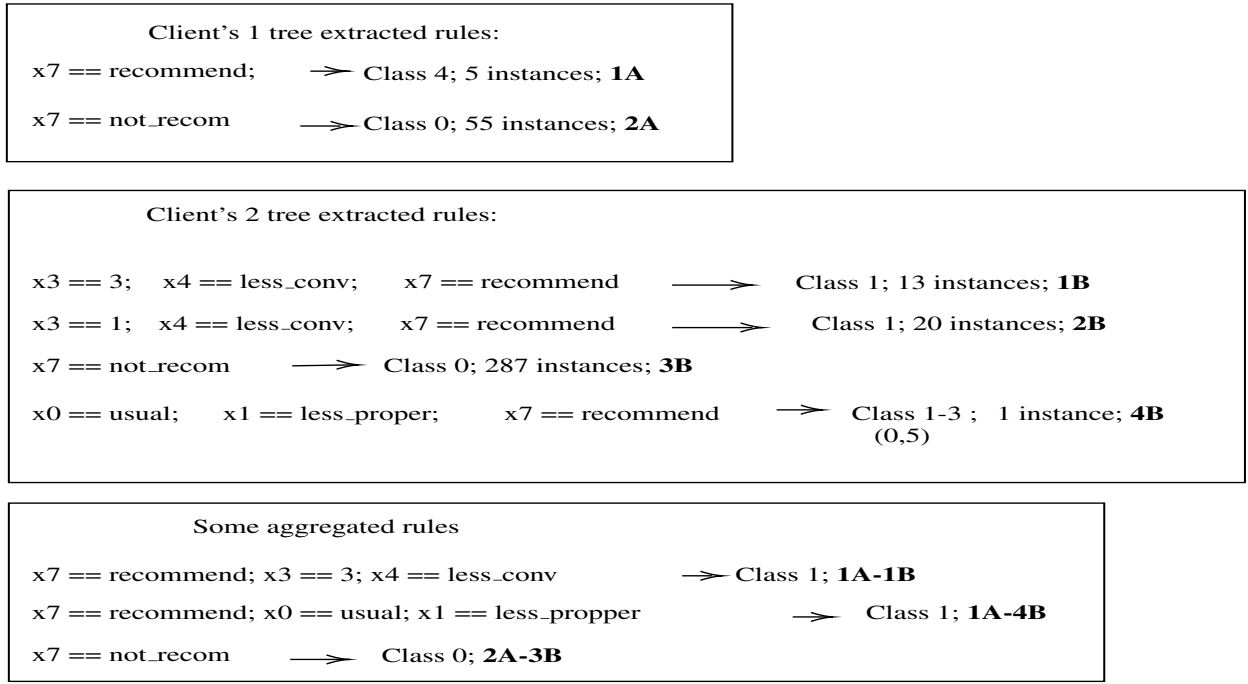
\begin{figure}[h!]

\tikzset{every picture/.style={line width=0.75pt}} %set default line width to 0.75pt 

\resizebox{\textwidth}{9cm}{ %new

\begin{tikzpicture}[x=0.75pt,y=0.75pt,yscale=-1,xscale=1]
%uncomment if require: \path (0,397); %set diagram left start at 0, and has height of 397

%Right Arrow [id:dp0022382172212154217] 
% \draw   (74,76.25) -- (87.2,76.25) -- (87.2,72) -- (96,80.5) -- (87.2,89) -- (87.2,84.75) -- (74,84.75) -- cycle ;
%Right Arrow [id:dp017304360381646955] 
% \draw   (74,112.25) -- (87.2,112.25) -- (87.2,108) -- (96,116.5) -- (87.2,125) -- (87.2,120.75) -- (74,120.75) -- cycle ;
%Shape: Rectangle [id:dp3974261211789809] 
\draw   (39,26) -- (400,26) -- (400,141) -- (39,141) -- cycle ;
\draw   (40,162) -- (655,162) -- (655,375) -- (40,375) -- cycle ;

%Shape: Rectangle [id:dp2060436428570904323] 
\draw   (40,390) -- (585,390) -- (585,520) -- (40,520) -- cycle ;

% Text Node
\draw (50,68) node [anchor=north west][inner sep=0.75pt]   [align=left] {x7 == recommend;};
% Text Node
% \draw (103,69) node [anchor=north west][inner sep=0.75pt]   [align=left] {recommend};
% Text Node
\draw (215,70) node [anchor=north west][inner sep=0.75pt]   [align=left] {Class 4; 5 instances; \textbf{1A}};
% Text Node
\draw (50,104) node [anchor=north west][inner sep=0.75pt]   [align=left] {x7 == not\_recom};
% Text Node
% \draw (103,105) node [anchor=north west][inner sep=0.75pt]   [align=left] {not\_recom};
% Text Node
\draw (215,106) node [anchor=north west][inner sep=0.75pt]   [align=left] {Class 0; 55 instances; \textbf{2A}};
% Text Node
\draw (108,40) node [anchor=north west][inner sep=0.75pt]   [align=left] {Client's 1 tree extracted rules:};
% Text Node
\draw (107,173) node [anchor=north west][inner sep=0.75pt]   [align=left] {Client's 2 tree extracted rules:};
% Text Node
\draw (50,222) node [anchor=north west][inner sep=0.75pt]   [align=left] {x3 == 3;};
% Text Node
% \draw (109,221) node [anchor=north west][inner sep=0.75pt]   [align=left] {3;};
% Text Node
\draw (115,222) node [anchor=north west][inner sep=0.75pt]   [align=left] {x4 == less\_conv;};
% Text Node
% \draw (183,219) node [anchor=north west][inner sep=0.75pt]   [align=left] {less\_conv;};
% Text Node
\draw (235,222) node [anchor=north west][inner sep=0.75pt]   [align=left] {x7 == recommend};
% Text Node
% \draw (328,219) node [anchor=north west][inner sep=0.75pt]   [align=left] {recommend};
% Text Node
\draw (430,222) node [anchor=north west][inner sep=0.75pt]   [align=left] {Class 1; 13 instances; \textbf{1B}};
% Text Node
\draw (50,252) node [anchor=north west][inner sep=0.75pt]   [align=left] {x3 == 1;};
% Text Node
% \draw (111,251) node [anchor=north west][inner sep=0.75pt]   [align=left] {1;};
% Text Node
\draw (113,252) node [anchor=north west][inner sep=0.75pt]   [align=left] {x4 == less\_conv;};
% Text Node
% \draw (185,250) node [anchor=north west][inner sep=0.75pt]   [align=left] {less\_conv;};
% Text Node
\draw (233,252) node [anchor=north west][inner sep=0.75pt]   [align=left] {x7 == recommend};
% Text Node
% \draw (330,250) node [anchor=north west][inner sep=0.75pt]   [align=left] {recommend};
% Text Node
\draw (438,252) node [anchor=north west][inner sep=0.75pt]   [align=left] {Class 1; 20 instances; \textbf{2B}};
% Text Node
\draw (50,283) node [anchor=north west][inner sep=0.75pt]   [align=left] {x7 == not\_recom};
% Text Node
% \draw (113,283) node [anchor=north west][inner sep=0.75pt]   [align=left] {not\_recom};
% Text Node
\draw (225,283) node [anchor=north west][inner sep=0.75pt]   [align=left] {Class 0; 287 instances; \textbf{3B}};
% Text Node
\draw (50,318) node [anchor=north west][inner sep=0.75pt]   [align=left] {x0 == usual;};
% Text Node
% \draw (113,317) node [anchor=north west][inner sep=0.75pt]   [align=left] {usual;};
% Text Node
\draw (145,318) node [anchor=north west][inner sep=0.75pt]   [align=left] {x1 == less\_proper;};
% Text Node
% \draw (216,316) node [anchor=north west][inner sep=0.75pt]   [align=left] {less\_proper;};
% Text Node
\draw (285,318) node [anchor=north west][inner sep=0.75pt]   [align=left] {x7 == recommend};
% Text Node
% \draw (361,316) node [anchor=north west][inner sep=0.75pt]   [align=left] {recommend};
% Text Node
\draw (458,318) node [anchor=north west][inner sep=0.75pt]   [align=left] {Class 1-3 ;\\ \ \ \ (0,5)};
% Text Node
% \draw (281,71) node [anchor=north west][inner sep=0.75pt]   [align=left] {5 instances};
% Text Node
% \draw (279,106) node [anchor=north west][inner sep=0.75pt]   [align=left] {55 instances};
% Text Node
% \draw (508,221) node [anchor=north west][inner sep=0.75pt]   [align=left] {13 instances};
% Text Node
% \draw (510,249) node [anchor=north west][inner sep=0.75pt]   [align=left] {20 instances};
% Text Node
% \draw (301,285) node [anchor=north west][inner sep=0.75pt]   [align=left] {287 instances};
% Text Node
\draw (530,318) node [anchor=north west][inner sep=0.75pt]   [align=left] {1 instance; \textbf{4B}};
% Connection Rule 1A
\draw    (190,75) -- (210,75) ;
\draw [shift={(208,75)}, rotate = 180.57] [color={rgb, 255:red, 0; green, 0; blue, 0 }  ][line width=0.75]    (10.93,-3.29) .. controls (6.95,-1.4) and (3.31,-0.3) .. (0,0) .. controls (3.31,0.3) and (6.95,1.4) .. (10.93,3.29)   ;
% Connection Rule 2A
\draw    (185,115) -- (205,115) ;
\draw [shift={(211,115)}, rotate = 180.55] [color={rgb, 255:red, 0; green, 0; blue, 0 }  ][line width=0.75]    (10.93,-3.29) .. controls (6.95,-1.4) and (3.31,-0.3) .. (0,0) .. controls (3.31,0.3) and (6.95,1.4) .. (10.93,3.29)   ;
% Connection Rule 1B
\draw    (370,231) -- (410,231) ;
\draw [shift={(410,231)}, rotate = 180.6] [color={rgb, 255:red, 0; green, 0; blue, 0 }  ][line width=0.75]    (10.93,-3.29) .. controls (6.95,-1.4) and (3.31,-0.3) .. (0,0) .. controls (3.31,0.3) and (6.95,1.4) .. (10.93,3.29)   ;
% Connection Rule 2B
\draw    (370,262) -- (410,262) ;
\draw [shift={(415,262)}, rotate = 180.6] [color={rgb, 255:red, 0; green, 0; blue, 0 }  ][line width=0.75]    (10.93,-3.29) .. controls (6.95,-1.4) and (3.31,-0.3) .. (0,0) .. controls (3.31,0.3) and (6.95,1.4) .. (10.93,3.29)   ;
% Connection Rule 3B
\draw    (181,289.88) -- (212,289.18) ;
\draw [shift={(214,289.2)}, rotate = 180.6] [color={rgb, 255:red, 0; green, 0; blue, 0 }  ][line width=0.75]    (10.93,-3.29) .. controls (6.95,-1.4) and (3.31,-0.3) .. (0,0) .. controls (3.31,0.3) and (6.95,1.4) .. (10.93,3.29)   ;
% Connection Rule 4B
\draw    (422,322) -- (442,322) ;
\draw [shift={(445,322)}, rotate = 180.6] [color={rgb, 255:red, 0; green, 0; blue, 0 }  ][line width=0.75]    (10.93,-3.29) .. controls (6.95,-1.4) and (3.31,-0.3) .. (0,0) .. controls (3.31,0.3) and (6.95,1.4) .. (10.93,3.29)   ;

%Straight Lines [id:da8459697169811838] % Connection 1A-1B
\draw    (375,443) -- (390,443) ;
\draw [shift={(395,443)}, rotate = 180] [color={rgb, 255:red, 0; green, 0; blue, 0 }  ][line width=0.75]    (10.93,-3.29) .. controls (6.95,-1.4) and (3.31,-0.3) .. (0,0) .. controls (3.31,0.3) and (6.95,1.4) .. (10.93,3.29)   ;

%Straight Lines [id:da8459697169811839] % Connection 1A-4B
\draw    (430,473) -- (450,473) ;
\draw [shift={(450,473)}, rotate = 180] [color={rgb, 255:red, 0; green, 0; blue, 0 }  ][line width=0.75]    (10.93,-3.29) .. controls (6.95,-1.4) and (3.31,-0.3) .. (0,0) .. controls (3.31,0.3) and (6.95,1.4) .. (10.93,3.29)   ;

%Straight Lines [id:da8459697169811840] % Connection 2A-3B
\draw    (180,504) -- (210,504) ;
\draw [shift={(209,504)}, rotate = 180] [color={rgb, 255:red, 0; green, 0; blue, 0 }  ][line width=0.75]    (10.93,-3.29) .. controls (6.95,-1.4) and (3.31,-0.3) .. (0,0) .. controls (3.31,0.3) and (6.95,1.4) .. (10.93,3.29)   ;

% Text Node
\draw (130,400) node [anchor=north west][inner sep=0.75pt]   [align=left] {Some aggregated rules};
% Text Node
\draw (50,435) node [anchor=north west][inner sep=0.75pt]   [align=left] {x7 == recommend; x3 == 3; x4 == less\_conv};
% Text Node
\draw (400,435) node [anchor=north west][inner sep=0.75pt]   [align=left] {Class 1; \textbf{1A-1B}};
% Text Node
\draw (50,463) node [anchor=north west][inner sep=0.75pt]   [align=left] {x7 == recommend; x0 == usual; x1 == less\_propper};
% Text Node
\draw (465,463) node [anchor=north west][inner sep=0.75pt]   [align=left] {Class 1; \textbf{1A-4B}};
% Text Node
\draw (50,494) node [anchor=north west][inner sep=0.75pt]   [align=left] {x7 == not\_recom};
% Text Node
\draw (224,494) node [anchor=north west][inner sep=0.75pt]   [align=left] {Class 0; \textbf{2A-3B}};

\end{tikzpicture}
} %new
\caption{Example of rules generated from the clients' trees when they build an ID3. Client 1 rules are obtained by decomposing the Client's 1 tree into rules, while Client 2 rules are only some rules generated by that client. In the figure appears the condition of the rules, the class predicted by the rule, the number of instances that fit the rule in the client's data, and, in bold, the name given to the rule for a better understanding of the aggregation method.}

% \caption{Ejemplo de reglas de asociación generadas en cada cliente cuando se utiliza como árbol base un ID3. Las reglas del cliente 1 son todas las que se han generado en ese cliente, mientras que las reglas del cliente 2 son solo algunas de las múltiples reglas generadas en este cliente.}
\label{fig:rules_example_id3}
\end{figure}

The $Server$ has a rule set for each local DT. For example, Figure \ref{fig:rules_example_id3} presents the rule set associated with certain clients $C_1$ and $C_2$. In this simulated experimental example, the first client has two rules while the second client has four rules. This step aggregates the compatible rules, \textit{i.e.}, we merge two rules when they do not contradict each other.
%Once the $Server$ has extracted the rule set for each decision tree, Figure \ref{fig:rules_example_id3}, he begins the aggregating process. To merge two rule sets, they have to be compatible, i.e., they don't contradict each other.\\ 

To merge two association rules using the cartesian product from two ID3 trees, A, and B, we need to ensure that no conditions on the antecedent of A are present in the conditions on the antecedent of B. This is because if two rules have overlapping antecedents, then they are essentially denoting the same thing. An example of this condition of aggregating rules can be seen in Figure \ref{fig:rules_example_id3} when trying to merge the rules from Client $C_1$ with the rules from Client $C_2$. First, we start aggregating the rule \textbf{1A}, and we start checking the compatibility with rule \textbf{1B}. Both rules, 1A and 1B, share the condition \textbf{\textit{x7==recommend}} on the antecedent, thus, they are compatible. The result of merging these rules is the rule \textbf{1A-1B}, i.e.,  \textbf{\textit{x7==recommend; x3==3; x4==less\_conv} $\rightarrow$ Class 1}. Afterward, the probability distribution of the classes for each rule is added to the new rule.\\

One example of rules that contradict each other is if we try to merge rules \textbf{2A} and \textbf{4B}. Rule \textbf{2A} has the condition \textbf{\textit{x7==not recommend}} on the antecedent, while rule \textbf{4B} has the condition \textbf{\textit{x7==recommend}} on the antecedent being, therefore, rules that contradict each other.

\paragraph{Build the global decision tree}

% The rule-set aggregated by the server is used to build a decision tree and this rule-set has the rules and the class probabilities. The tree's structure is based on the tree algorithm used by the clients and here, it has an ID3 architecture. As we are building an ID3 tree, we stop the building process when there isn't any feature left, as we use the features once a time in each branch.

The Server uses the aggregated rule set to build an ID3 DT. The tree-building process stops when no more features are left, or when the conditions presented in Section \ref{subsec:aggmethod} are met. % as our ID3 implementation uses each feature only once. %We use the Information Gain (IG) \cite{SAGI2020124} to select each feature $f_{i} \in F$ as shown in Equation \ref{eq:informationgaingeneral}.
We compute the Information Gain (IG) for each DT node and select the feature $f_{r} \in F$ with the higher value of IG to split the node. The IG value associated to the feature $f_{r}$ is computed as follows:
\begin{equation}\label{eq:informationgaingeneral}
    IG (RS_{r}, R) = \frac{\sum_{s=1}^{k}(|RS_{rs}|*\textit{entropy}(RS_{rs}))}{\sum_{s=1}^{k}|RS_{rs}|} -\textit{entropy}(RS_{r}) 
\end{equation}
such as the entropy of the node is calculated as the entropy of ($argmax(\hat{y}_1$), $argmax(\hat{y}_2$), $argmax(\hat{y}_K$)) of all of the conjunctions included in the node \cite{SAGI2020124}. 

\subsection{CART rules aggregation and tree building (ICDTA4FL-CART model)}\label{subsec:cartagg}

In this section, we present the steps needed for the ICDTA4FL process to work on a CART DT model, \textit{i.e.}, all the clients train locally in a CART, namely the ICDTA4FL-CART model. Next, we describe the two-dependant steps in the construction of the global DT.\\

% There are two steps that are dependent on the tree: the aggregation of the rules and the construction of the global decision tree using these aggregated rules.

\paragraph{Aggregate the rules}

The contradict condition on a CART rule set differs from the ID3, as the conditions represent intervals from the binary split, Figure \ref{fig:rules_example_cart}. In this case, if two rules share a condition, \textit{i.e.}, rule \textbf{2A}, and rule \textbf{4B}, we select the less restrictive condition. While rule \textbf{2A} has the condition \textbf{\textit{$x0 > 32.5$}}, the rule \textbf{4B} has the condition \textbf{\textit{$x0 > 35$}}, being the second condition more restrictive than the first one, so when merging the rules we keep the condition from rule \textbf{2A}.\\
\begin{figure}[h!]

\tikzset{every picture/.style={line width=0.75pt}} %set default line width to 0.75pt        

\resizebox{\textwidth}{8cm}{ %new

\begin{tikzpicture}[x=0.75pt,y=0.75pt,yscale=-1,xscale=1]
%uncomment if require: \path (0,528); %set diagram left start at 0, and has height of 528

%Shape: Rectangle [id:dp8633007252014324] % Client  1
\draw   (31,20) -- (332.5,20) -- (332.5,113) -- (31,113) -- cycle ;
%Shape: Rectangle [id:dp14117132509368857] % Client 2
\draw   (32,128) -- (600,128) -- (600,313) -- (32,313) -- cycle ;
%Straight Lines [id:da3276682764479939] 
\draw    (133,65) -- (150.5,65) ;
\draw [shift={(152.5,65)}, rotate = 180] [color={rgb, 255:red, 0; green, 0; blue, 0 }  ][line width=0.75]    (10.93,-3.29) .. controls (6.95,-1.4) and (3.31,-0.3) .. (0,0) .. controls (3.31,0.3) and (6.95,1.4) .. (10.93,3.29)   ;
%Straight Lines [id:da6437877351955253] 
\draw    (122,91) -- (139.5,91) ;
\draw [shift={(141.5,91)}, rotate = 180] [color={rgb, 255:red, 0; green, 0; blue, 0 }  ][line width=0.75]    (10.93,-3.29) .. controls (6.95,-1.4) and (3.31,-0.3) .. (0,0) .. controls (3.31,0.3) and (6.95,1.4) .. (10.93,3.29)   ;
%Straight Lines [id:da0312752252420162] % Connection 1B
\draw    (372,175) -- (398.5,175) ;
\draw [shift={(399.5,175)}, rotate = 180] [color={rgb, 255:red, 0; green, 0; blue, 0 }  ][line width=0.75]    (10.93,-3.29) .. controls (6.95,-1.4) and (3.31,-0.3) .. (0,0) .. controls (3.31,0.3) and (6.95,1.4) .. (10.93,3.29)   ;
%Straight Lines [id:da33614615094014755] % Connection 2B
\draw    (357,200) -- (374.5,200) ;
\draw [shift={(376.5,200)}, rotate = 180] [color={rgb, 255:red, 0; green, 0; blue, 0 }  ][line width=0.75]    (10.93,-3.29) .. controls (6.95,-1.4) and (3.31,-0.3) .. (0,0) .. controls (3.31,0.3) and (6.95,1.4) .. (10.93,3.29)   ;
%Straight Lines [id:da6373199544086117] % Connection 3B
\draw    (294,228) -- (311.5,228) ;
\draw [shift={(313.5,228)}, rotate = 180] [color={rgb, 255:red, 0; green, 0; blue, 0 }  ][line width=0.75]    (10.93,-3.29) .. controls (6.95,-1.4) and (3.31,-0.3) .. (0,0) .. controls (3.31,0.3) and (6.95,1.4) .. (10.93,3.29)   ;
%Straight Lines [id:da514192388224315] % Connection 4B
\draw    (360,254) -- (384.5,254) ;
\draw [shift={(386.5,254)}, rotate = 180] [color={rgb, 255:red, 0; green, 0; blue, 0 }  ][line width=0.75]    (10.93,-3.29) .. controls (6.95,-1.4) and (3.31,-0.3) .. (0,0) .. controls (3.31,0.3) and (6.95,1.4) .. (10.93,3.29)   ;
%Straight Lines [id:da3240174934367116] % Connection 5B
\draw    (348,286) -- (365.5,286) ;
\draw [shift={(367.5,286)}, rotate = 180] [color={rgb, 255:red, 0; green, 0; blue, 0 }  ][line width=0.75]    (10.93,-3.29) .. controls (6.95,-1.4) and (3.31,-0.3) .. (0,0) .. controls (3.31,0.3) and (6.95,1.4) .. (10.93,3.29)   ;

%Shape: Rectangle [id:dp08421316449940663] % Some aggregated rules
\draw   (33,330) -- (541.5,330) -- (541.5,479) -- (33,479) -- cycle ;
%Straight Lines [id:da4331372990045672] 
\draw    (315,380) -- (346.5,380) ;
\draw [shift={(348.5,380)}, rotate = 180] [color={rgb, 255:red, 0; green, 0; blue, 0 }  ][line width=0.75]    (10.93,-3.29) .. controls (6.95,-1.4) and (3.31,-0.3) .. (0,0) .. controls (3.31,0.3) and (6.95,1.4) .. (10.93,3.29)   ;
%Straight Lines [id:da20061343865836212] 
\draw    (373,411) -- (405.5,411) ;
\draw [shift={(407.5,411)}, rotate = 180] [color={rgb, 255:red, 0; green, 0; blue, 0 }  ][line width=0.75]    (10.93,-3.29) .. controls (6.95,-1.4) and (3.31,-0.3) .. (0,0) .. controls (3.31,0.3) and (6.95,1.4) .. (10.93,3.29)   ;
%Straight Lines [id:da8459697169811838] 
\draw    (360, 440) -- (394.5,440) ;
\draw [shift={(396.5,440)}, rotate = 180] [color={rgb, 255:red, 0; green, 0; blue, 0 }  ][line width=0.75]    (10.93,-3.29) .. controls (6.95,-1.4) and (3.31,-0.3) .. (0,0) .. controls (3.31,0.3) and (6.95,1.4) .. (10.93,3.29)   ;

% Text Node
\draw (123,28) node [anchor=north west][inner sep=0.75pt]   [align=left] {Client 1 extracted rules:};
% Text Node
\draw (50,55) node [anchor=north west][inner sep=0.75pt]   [align=left] {x0 $<=$ 32.5 };
% Text Node
\draw (153,55) node [anchor=north west][inner sep=0.75pt]   [align=left] {Class 1; 2 instances; \textbf{1A}};
% Text Node
\draw (50,82) node [anchor=north west][inner sep=0.75pt]   [align=left] {x0 $>$ 32.5 };
% Text Node
\draw (143,82) node [anchor=north west][inner sep=0.75pt]   [align=left] {Class 2; 4 instances; \textbf{2A}};
% Text Node
\draw (233,137) node [anchor=north west][inner sep=0.75pt]   [align=left] {Client 2 extracted rules:};
% Text Node
\draw (50,166) node [anchor=north west][inner sep=0.75pt]   [align=left] {x3$>$49; x54$<$=197085; x64$<=$0.5; x71$<=$14.5};
% Text Node
\draw (401,166) node [anchor=north west][inner sep=0.75pt]   [align=left] {Class 1; 105 instances; \textbf{1B}};
% Text Node
\draw (50,192) node [anchor=north west][inner sep=0.75pt]   [align=left] {x3$>$49; x54$>$197085; x64$<=$0.5; x71$<=$14.5};
% Text Node
\draw (384,191) node [anchor=north west][inner sep=0.75pt]   [align=left] {Class 1; 1 instances; \textbf{2B}};
% Text Node
\draw (50,219) node [anchor=north west][inner sep=0.75pt]   [align=left] {x0$>$39; x74$>$0.5;x3$>$22;x71$<$=10.5};
% Text Node
\draw (321,219) node [anchor=north west][inner sep=0.75pt]   [align=left] {Class 2; 31 instances; \textbf{3B}};
% Text Node
\draw (50,247) node [anchor=north west][inner sep=0.75pt]   [align=left] {x0$>$35; x74$>$0.5;x3$>$22;x71$>$10.5; x51$<=$0.5};
% Text Node
\draw (394,245) node [anchor=north west][inner sep=0.75pt]   [align=left] {Class 2; 29 instances; \textbf{4B}};
% Text Node
\draw (50,278) node [anchor=north west][inner sep=0.75pt]   [align=left] {x0$>$35; x74$>$0.5;x3$>$22;x71$>$10.5; x51$>$0.5};
% Text Node
\draw (375,276) node [anchor=north west][inner sep=0.75pt]   [align=left] {Class 1; 3 instances; \textbf{5B}};
% Text Node
\draw (138,339) node [anchor=north west][inner sep=0.75pt]   [align=left] {Some aggregated rules};
% Text Node
\draw (50,375) node [anchor=north west][inner sep=0.75pt]   [align=left] {x0$>$32.5; x74$>$0.5;x3$>$22;x71$<=$10.5};
% Text Node
\draw (356,375) node [anchor=north west][inner sep=0.75pt]   [align=left] {Class 2; \textbf{2A-3B}};
% Text Node
\draw (50,403) node [anchor=north west][inner sep=0.75pt]   [align=left] {x0$>$32.5; x74$>$0.5;x3$>$22;x71$>$10.5; x51$<=$0.5};
% Text Node
\draw (415,402) node [anchor=north west][inner sep=0.75pt]   [align=left] {Class 2; \textbf{2A-4B}};
% Text Node
\draw (50,434) node [anchor=north west][inner sep=0.75pt]   [align=left] {x0$>$32.5; x74$>$0.5;x3$>$22;x71$>$10.5; x51$>$0.5};
% Text Node
\draw (404,433) node [anchor=north west][inner sep=0.75pt]   [align=left] {Class 2; \textbf{2A-5B}};

\end{tikzpicture}

} %new

\caption{Example of rules generated from the clients' trees when they have built a CART. Client 1 rules are obtained by decomposing the Client's 1 tree into rules, while Client 2 rules are only some rules generated by that client. In the figure appears the condition of the rules, the class predicted by the rule, the number of instances that fit the rule in the client's data, and, in bold, the name given to the rule for a better understanding of the aggregation method.
}

% \caption{Ejemplo de reglas de asociación generadas en cada cliente cuando se utiliza como árbol base un CART. Las reglas del cliente 1 son todas las que se han generado en ese cliente, mientras que las reglas del cliente 2 son solo algunas de las múltiples reglas generadas en este cliente.}
\label{fig:rules_example_cart}
\end{figure}

Examples of rules contradicting each other are the rules \textbf{1A} and rule \textbf{3B}. The rule \textbf{1A} has the condition \textbf{\textit{$x0 <= 32.5$}}, while the rule \textbf{3B} has the condition \textbf{\textit{$x0 > 35$}}, making them incompatible.

\paragraph{Building the global decision tree}

A CART-tree type is a binary tree, so the metric to split a node is given by the Equation \ref{eq:informationgainbinary}. The conditions to stop building the tree are described in Section \ref{subsec:aggmethod}.
\begin{footnotesize}
    \begin{equation}\label{eq:informationgainbinary}
        IG(RS_{i}, R) = \frac{|RS_{i1}|*\textit{entropy}(RS_{i1}) + |RS_{i2}|*\textit{entropy}(RS_{i2}) }{|RS_{i1}| + |RS_{i2}|} -\textit{entropy}(RS_{i})
    \end{equation}
\end{footnotesize}

%% SECTION 4 - EXPERIMENTAL FRAMEWORK
\section{Experimental Framework}\label{sec:expframe}

This section describes the experimental framework.  We pay attention to the following four elements in the following sections:\begin{enumerate*}[label=(\arabic*)]
    \item Data and data distribution,
    \item Metrics and clients,
    \item Decision tree parameters, and
    \item Comparison with the state-of-the-art.
\end{enumerate*}

\subsection{Data and Data Distribution}\label{subsec:efdata}

We have selected 4 classification datasets to conduct the experimental study: Nursery\footnote{\url{https://archive.ics.uci.edu/ml/datasets/nursery}},
Adult\footnote{\url{https://archive.ics.uci.edu/ml/datasets/adult}}, Car\footnote{\url{https://archive.ics.uci.edu/dataset/19/car+evaluation}}, and Credit2\footnote{\url{https://archive.ics.uci.edu/ml/datasets/default+of+credit+card+clients}}. The information related to the datasets is shown in Table \ref{tab:basesdedatos}.

\begin{table}[!h]
    \centering
    \begin{tabularx}{\linewidth}{
    @{}Xr@{\hphantom{1}}r@{\hphantom{1}}r@{\hphantom{1}}r@{}
    }
    \toprule
    \textbf{Name} & \textbf{Dataset type} & \textbf{Instances} & \textbf{Features} & \textbf{Classes} \\ \midrule
    \textbf{Adult} & Numerical and Categorical & 48842 & 14 & 2 \\
    \textbf{Nursery} & Categorical & 12960 & 8 & 5 \\
    \textbf{Car} & Categorical & 1728 & 6 & 4 \\
    \textbf{Credit} & Numerical & 30000 & 24 & 2 \\ 
    \bottomrule
    \end{tabularx}
    \caption{Datasets used in the experimental framework.}
    \label{tab:basesdedatos}
\end{table}

Data distribution is very important when working in federated scenarios, where we can find two distributions: IID and non-IID. We use both IID and non-IID data distributions to evaluate the performance of the models. 
\begin{itemize}
    \item In the IID distribution, all clients have the same classes and the same number of instances.
    \item In the non-IID distribution, clients can have different numbers of instances and different numbers of classes.
\end{itemize}  
We randomize the partitions made in the non-IID distribution to ensure that each client has at least 5 instances.\\

\subsection{Metrics and Clients}\label{subsec:efmetrics}

We consider the accuracy (Acc) and the Macro-F1 (F1) metrics to evaluate the performance of the ICDTA4FL model. As it is essential to maintain the performance of the trained model while increasing the number of clients, it may vary depending on the task being resolved, or the clients available for training the model. In our experiments, we use 2, 10, 20, and 50 as the number of clients to analyze their performance when going to large-scale scenarios.\\

We conduct a 10-fold cross-validation (FCV) at the client level and compute the mean value to obtain the results. Each client evaluates his/her local DT and the global tree with its local test set. To get the results, we average the 10 FCV across each client and then average the results for each client to get the final model metrics. \\

\subsection{Decision tree parameters} \label{subsec:efdtp}

As shown in Section \ref{sec:proposal}, the ICDTA4FL process works with different types of DTs, and, to prove so, we use ID3 and CART as base trees for the local models trained by the clients. We set 2 parameters when creating each DT on the client side: the max\_depth and the criterion. For the CART tree, we set the max\_depth to 5 and the gini-index as the criterion, while the ID3 uses a $\dfrac{|F|}{2}$ as max\_depth and information gain as the criterion. Here $F$ represents the feature set of the dataset. Therefore, we use the same max\_depth for building the global model, and the criteria defined in Section \ref{sec:proposal}. Equation \ref{eq:informationgaingeneral} servers as the criterion for the global model when building an ID3, and Equation \ref{eq:informationgainbinary} servers as the criterion when building a CART. \\

\subsection{Federated-ID3 model: A federated learning based on ID3 as state-of-the-art for comparison} \label{subsec:sota}

We showed in Section \ref{sec:background} that the state-of-the-art models build ensembles of trees, following bagging or boosting strategies, but no work builds a single DT as ICDTA4FL does. The Federated-ID3 model \cite{truex2019hybrid} is the only method that builds a single DT rather than an ensemble of trees, so we use this model to compare with the ICDTA4FL-ID3 model. In Federated-ID3 \cite{truex2019hybrid}, the algorithm selects the feature $F$ that maximizes the information gain. It proceeds until reaching a leaf node either when no more features are available or when it reaches the maximum depth, $\dfrac{|F|}{2}$, being $F$ the feature set of the dataset. Note that the Federated-ID3 model builds a single DT aggregating the counts and class counts to calculate the information gain, while the ICDTA4FL-ID3 model builds a global DT based on the local DTs of the clients.\\

%% SECTION 5 - EXPERIMENTAL RESULTS AND ANALYSIS
\section{Experimental Results and Analysis}\label{sec:exresanlysis}
% In this section wtih presente the experimetental results for ooour study with both models (ID3 and CART) and a complete and Deep  analysis. It is organized as follow:  Section 5.1., ...
In this section we present the experimental results for our study with both models (ID3 and CART) and a complete and deep analysis. It is organized as follows: Section \ref{subsec:resultsID3} shows the results of ICDTA4FL using ID3, that is the ICDTA4FL-ID3 model, against the state-of-the-art, the Federated-ID3 model \cite{truex2019hybrid}. Section \ref{subsec:resultsCART} shows the results obtained by ICDTA4FL using CART as the base model, that is the ICDTA4FL-CART model, and compares it to the ICDTA4FL-ID3 model. Section \ref{subsec:ernc} analyses the performance of the ICDTA4FL-ID3 and CART models when varying the number of clients. Section \ref{subsec:filter} shows the robustness of the ICDTA4FL process when using the percentile as the filter method for selecting clients. Lastly, Section \ref{subsec:interpretability} illustrates the interpretability of the ICDTA4FL process over the local DT built by the clients.

\subsection{The ICDTA4FL-ID3 model: Results and Analysis}\label{subsec:resultsID3}

In this section, we present the results obtained by the proposed ICDTA4FL process when clients train an ID3 locally. We compare the results of the ICDTA4FL-ID3 model against the state-of-the-art, the Federated-ID3 model \cite{truex2019hybrid}. We present two tables, Tables \ref{tab:id3comparacioniid} and \ref{tab:id3comparacionnoiid}, in which we show the results of the ICDTA4FL-ID3 model compared with the state-of-the-art for the datasets presented in Section \ref{subsec:efdata}. Table \ref{tab:id3comparacioniid} presents the results when there is an IID data distribution, while Table \ref{tab:id3comparacionnoiid} presents the results when there is a non-IID data distribution. The baseline model presents the mean result of the local DTs evaluated on its associated client. \\

The datasets Adult and Credit2 are not fully categorical. We conduct a basic preprocessing step to transform all the features into categorical so that ICDTA4FL-ID3 and the Federated-ID3 model \cite{truex2019hybrid} can build the DTs. \\

%\vspace{-.5cm}
\begin{table}[!bh]
    \centering
    \begin{footnotesize}
    \begin{tabularx}{\linewidth}{   @{}r@{\hphantom{10}}X@{\hphantom{10}}r@{\hphantom{10}}r@{\hphantom{10}}r@{\hphantom{10}}r@{\hphantom{10}}r@{\hphantom{10}}r@{\hphantom{10}}r@{\hphantom{10}}r@{}
    }
    \toprule 
    % Cabecera
    &  & \multicolumn{2}{c}{\textbf{Nursery}} & \multicolumn{2}{c}{\textbf{Adult}} & \multicolumn{2}{c}{\textbf{Car}} & \multicolumn{2}{c}{\textbf{Credit2}}\\ \midrule
    & & \textbf{Acc} & \textbf{F1} & \textbf{Acc} & \textbf{F1} & \textbf{Acc} & \textbf{F1} & \textbf{Acc} & \textbf{F1} \\ \midrule
    % 2 Clients
    \multirow{3}{*}{\rotatebox{90}{2 Clients}} & \textbf{Baseline (ID3) } & 90.83 & 69.85 & 81.37 & 73.31 & 88.01 & 65.43 & 80.22 & \textbf{66.21}  \\
    \cmidrule{2-10}
    & \textbf{ICDTA4FL-ID3} & \textbf{90.93} & \textbf{73.75} & 81.53 & \textbf{73.64} & \textbf{90.08} & \textbf{74.6} & 80.35 & 65.98  \\
    & \textbf{Federated-ID3 \cite{truex2019hybrid}} & 89.22 & 60.18 & \textbf{83.64} & 64.14 & 77.65 & 37.88 & \textbf{83.71} & 65.96  \\ 
    \midrule
    % 5 Clients
    \multirow{3}{*}{\rotatebox{90}{5 Clients}} & \textbf{Baseline (ID3)} & 90.41 & \textbf{78.26} & 80.87 & \textbf{73.13} & \textbf{82.44} & \textbf{56.19} & 81.48 & 67.61  \\ 
    \cmidrule{2-10}
    & \textbf{ICDTA4FL-ID3} & \textbf{91.75} & 76.22 & 80.52 & 72.96 & 79.59 & 43.57 & \textbf{82.81} & \textbf{68.39}  \\
    & \textbf{Federated-ID3 \cite{truex2019hybrid}} & 88.79 & 63.07 & \textbf{81.51} & 55.06 & 77.37 & 37.17 & 75.34 & 49.61  \\ 
    \midrule
    % 10 Clients
    \multirow{3}{*}{\rotatebox{90}{10 Clients}} & \textbf{Baseline (ID3)} & 88.65 & \textbf{70.56} & 88.18 & 72.39 & \textbf{78.33} & \textbf{49.45} & 77.27 & 63.01  \\ 
    \cmidrule{2-10}
    & \textbf{ICDTA4FL-ID3} & \textbf{89.85} & 68.31 & \textbf{89.97} & \textbf{72.48} & 77.91 & 48.79 & \textbf{81.1} & \textbf{67.17}  \\ 
    & \textbf{Federated-ID3 \cite{truex2019hybrid}} & 88.42 & 66.3 & 79.8 & 51.17 & 72.83 & 38.8 & 74.87 & 51.97  \\ 
    \midrule
    % 20 Clients
    \multirow{3}{*}{\rotatebox{90}{20 Clients}} & \textbf{Baseline (ID3)} & 85.19 & 63.81 & 77.96 & \textbf{68.85} & 75.41 & 49.86 & 78.18 & 63.55  \\ 
    % & 80.325 & 71.615 & 88.339 & 71.07 & 86.135 & 85.19 & 79.159 & 64.645 \\
    \cmidrule{2-10}
    & \textbf{ICDTA4FL-ID3} & 87.5 & \textbf{74.54} & \textbf{78.57} & 63.87 & \textbf{77.5} & \textbf{57.54} & \textbf{80.95} & \textbf{64.13}  \\ 
    % & 79.778 & \underline{64.383} & \textbf{89.513} & \textbf{70.487} & \underline{53.03} & \underline{34.277} & \textbf{81.7} & \textbf{65.848} \\
    & \textbf{Federated-ID3 \cite{truex2019hybrid}} & \textbf{88.12} & 66.46 & 77.03 & 47.81 & 72.67 & 36.44 & 65.30 & 56.25  \\ 
    % & \textbf{81.82} & 50.33 & \underline{89.15} & 67.11 & \textbf{91.86} & \textbf{89.751} & 74.817 & 41.708 \\
    \midrule
    % 50 Clients
    \multirow{3}{*}{\rotatebox{90}{50 Clients}} & \textbf{Baseline (ID3)} & 82.10 & 64.55 & 77.23 & 66.12 & 62.8 & 45.44 & 75.77 & 61.29  \\ 
    % & 77.972 & 68.698 & 82.553 & 64.04 & 80.59 & 78.30 & 79.79 & 62.57 \\
    \cmidrule{2-10}
    & \textbf{ICDTA4FL-ID3} & \textbf{87.42} & \textbf{74.04} & \textbf{81.8} & \textbf{70.41} & \textbf{67.6} & \textbf{53.55} & \textbf{81.31} & \textbf{64.93}  \\ 
    & \textbf{Federated-ID3 \cite{truex2019hybrid}} & 84.42 & 66.11 & 71.45 & 44.82 & 62.13 & 36.22 & 69.83 & 50.97  \\
    \bottomrule
    \end{tabularx}
    \caption{IID distribution results of the ICDTA4FL-ID3 model against the state-of-the-art, the Federated-ID3 model. Also, we compare the ICDTA4FL-ID3 model with the baseline model. The baseline model refers to the mean of the local trees built by the clients when using ID3 as the base model.}
    \label{tab:id3comparacioniid}
    \end{footnotesize}
\end{table}

% \setlength\dashlinedash{0.2pt}
% \setlength\dashlinegap{1.5pt}
% \setlength\arrayrulewidth{0.3pt}

%\vspace{-.5cm}
\begin{table}[!bh]
    \centering
    \begin{footnotesize}
    \begin{tabularx}{\linewidth}{
    @{}r@{\hphantom{10}}X@{\hphantom{10}}r@{\hphantom{10}}r@{\hphantom{10}}r@{\hphantom{10}}r@{\hphantom{10}}r@{\hphantom{10}}r@{\hphantom{10}}r@{\hphantom{10}}r@{}
    }
    \toprule
    % Cabecera
    &  & \multicolumn{2}{c}{\textbf{Nursery}} & \multicolumn{2}{c}{\textbf{Adult}} & \multicolumn{2}{c}{\textbf{Car}} & \multicolumn{2}{c}{\textbf{Credit2}}\\ \midrule
    & & \textbf{Acc} & \textbf{F1} & \textbf{Acc} & \textbf{F1} & \textbf{Acc} & \textbf{F1} & \textbf{Acc} & \textbf{F1} \\ \midrule
    % 2 Clients
    \multirow{3}{*}{\rotatebox{90}{2 Clients}} & \textbf{Baseline (ID3)} & 91.04 & 70.22 & 80.01 & 71.8 & 80.83 & 45.09 & 80.67 & 67.17  \\ 
    % & 81.904 & 73.491 & 90.87 & 69.452 & 87.322 & 86.323 & 81.096 & 66.809 \\ 
    \cmidrule{2-10}
    & \textbf{ICDTA4FL-ID3} & \textbf{91.45} & \textbf{71.45} & 81.66 & \textbf{73.36} & \textbf{86.25} & \textbf{60.93} & 80.96 & \textbf{67.55}  \\
    & \textbf{Federated-ID3 \cite{truex2019hybrid}} & 89.22 & 59.57 & \textbf{83.47} & 64.03 & 77.29 & 37.31 & \textbf{83.81} & 67.02  \\
    \midrule
    % 5 Clients
    \multirow{3}{*}{\rotatebox{90}{5 Clients}} & \textbf{Baseline (ID3)} & 90.39 & 73.52 & 79.5 & 71.45 & 79.61 & \textbf{54.37} & 78.81 & 62.96  \\
    \cmidrule{2-10}
    & \textbf{ICDTA4FL-ID3} & \textbf{91.75} & \textbf{75.31} & \textbf{82.45} & \textbf{73.81} & \textbf{79.95} & 45.76 & \textbf{80.4} & \textbf{64.7}  \\
    & \textbf{Federated-ID3 \cite{truex2019hybrid}} & 88.86 & 63.96 & 81.74 & 54.69 & 76.32 & 36.81 & 75.31 & 50.63  \\
    \midrule
    % 10 Clients
    \multirow{3}{*}{\rotatebox{90}{10 Clients}} & \textbf{Baseline (ID3)} & 86.46 & 70.62 & 77.47 & \textbf{69.33} & 75.41 & \textbf{55.03} & 73.94 & 60.03  \\
    \cmidrule{2-10}
    & \textbf{ICDTA4FL-ID3} & \textbf{91.52} & \textbf{75.08} & \textbf{80.39} & 68.75 & \textbf{79.74} & 54.02 & \textbf{84.28} & \textbf{71.53}  \\
    & \textbf{Federated-ID3 \cite{truex2019hybrid}} & 88.16 & 65.15 & 79.69 & 51.08 & 75.95 & 35.45 & 73.94 & 45.01  \\
    \midrule
    % 20 Clients
    \multirow{3}{*}{\rotatebox{90}{20 Clients}} & \textbf{Baseline (ID3)} & 85.96 & 65.92 & 76.47 & 66.97 & 66.94 & 42.44 & 75.58 & \textbf{59.48}  \\
    \cmidrule{2-10}
    & \textbf{ICDTA4FL-ID3} & \textbf{88.4} & \textbf{69.89} & \textbf{78.98} & \textbf{67.86} & \textbf{77.99} & \textbf{53.89} & \textbf{77.58} & 53.66  \\
    & \textbf{Federated-ID3 \cite{truex2019hybrid}} & 87.47 & 67.69 & 77.09 & 48.18 & 74.19 & 37.2 & 66.31 & 56.85  \\
    \midrule
    % 50 Clients
    \multirow{3}{*}{\rotatebox{90}{50 Clients}} & \textbf{Baseline (ID3)} & 82.34 & 64.56 & 76.77 & \textbf{64.13} & 62.48 & 45.56 & 73.56 & 58.16  \\
    \cmidrule{2-10}
    & \textbf{ICDTA4FL-ID3} & \textbf{85.35} & \textbf{72.45} & \textbf{78.04} & 52.43 & \textbf{70.64} & \textbf{49.44} & \textbf{81.05} & \textbf{63.8}  \\
    & \textbf{Federated-ID3 \cite{truex2019hybrid}} & 84.31 & 64.21 & 71.6 & 44.97 & 66.28 & 39.1 & 67.83 & 49.69  \\
    \bottomrule
    \end{tabularx}
    \caption{Non-IID distribution results of the ICDTA4FL-ID3 model against the state-of-the-art, the Federated-ID3 model. Also, we compare the ICDTA4FL-ID3 model with the baseline model. The baseline model refers to the mean of the local trees built by the clients when using ID3 as the base model.}
    \label{tab:id3comparacionnoiid}
    \end{footnotesize}
\end{table}

Tables \ref{tab:id3comparacioniid} and \ref{tab:id3comparacionnoiid} presents the results of ICDTA4FL-ID3 against the state-of-the-art, the Federated-ID3 model \cite{truex2019hybrid}. Both tables show that ICDTA4FL-ID3 improves the state-of-the-art model and improves the baseline model, i.e., the local models trained by each client using their local data.\\

\begin{table}[!bht]
    \centering
    \begin{footnotesize}
    \begin{tabularx}{\linewidth}{@{}r@{\hphantom{10}}X@{\hphantom{10}}r@{\hphantom{10}}r@{\hphantom{10}}r@{\hphantom{10}}r@{\hphantom{10}}r@{\hphantom{10}}r@{\hphantom{10}}r@{\hphantom{10}}r@{}
    }
    \toprule 
    % Cabecera
    &  & \multicolumn{2}{c}{\textbf{Nursery}} & \multicolumn{2}{c}{\textbf{Adult}} & \multicolumn{2}{c}{\textbf{Car}} & \multicolumn{2}{c}{\textbf{Credit2}}\\ \midrule
    & & \textbf{Acc} & \textbf{F1} & \textbf{Acc} & \textbf{F1} & \textbf{Acc} & \textbf{F1} & \textbf{Acc} & \textbf{F1} \\ \midrule
    % 2 Clients
    \multirow{3}{*}{\rotatebox{90}{2 Clients}} & \textbf{Baseline (CART)} & 87.65 & 65.13 & 84.74 & 76.48 & 85.04 & 55.21 & 81.73 & 66.88 \\
    \cmidrule{2-10}
    & \textbf{ICDTA4FL-ID3} & \textbf{90.93} & \textbf{73.75} & 81.53 & 73.64 & \textbf{90.08} & \textbf{74.6} & 80.35 & 65.98 \\
    & \textbf{ICDTA4FL-CART} & 87.87 & 65.3 & \textbf{84.94} & \textbf{76.96} & 86.04 & 55.91 & \textbf{81.9} & \textbf{67.45} \\ 
    \midrule
    % 5 Clients
    \multirow{3}{*}{\rotatebox{90}{5 Clients}} & \textbf{Baseline (CART)} & 87.09 & 65.56 & 84.51 & 76.57 & 83.51 & 56.63 & 81.55 & 67.32 \\
    \cmidrule{2-10}
    & \textbf{ICDTA4FL-ID3} & \textbf{91.75} & \textbf{76.22} & 80.52 & 72.96 & 79.59 & 43.57 & \textbf{82.81} & \textbf{68.39} \\
    & \textbf{ICDTA4FL-CART} & 89.2 & 67.17 & \textbf{85.01} & \textbf{77.32} & \textbf{87.75} & \textbf{63.79} & 81.96 & 67.82 \\ 
    \midrule
    % 10 Clients
    \multirow{3}{*}{\rotatebox{90}{10 Clients}} & \textbf{Baseline (CART)} & 86.51 & 65.01 & 83.76 & 75.47 & 82.01 & 54.94 & 80.43 & 65.36 \\
    \cmidrule{2-10}
    & \textbf{ICDTA4FL-ID3} & \textbf{89.85} & \textbf{68.31} & \textbf{89.97} & 72.48 & 77.91 & 48.79 & \textbf{81.1} & \textbf{67.17} \\ 
    & \textbf{ICDTA4FL-CART} & 88.37 & 66.37 & 84.65 & \textbf{76.81} & \textbf{87.04} & \textbf{63.24} & 81.08 & 65.24 \\
    \midrule
    % 20 Clients
    \multirow{3}{*}{\rotatebox{90}{20 Clients}} & \textbf{Baseline (CART)} & 85.34 & 65.27 & 83.19 & 74.86 & 68.19 & 36.24 & 79.74 & 64.82 \\ 
    \cmidrule{2-10}
    & \textbf{ICDTA4FL-ID3} & \textbf{87.5} & \textbf{74.54} & 78.57 & 63.87 & \textbf{77.5} & \textbf{57.54} & \textbf{80.95} & 64.13 \\ 
    & \textbf{ICDTA4FL-CART} & 87.1 & 66.67 & \textbf{84.39} & \textbf{76.64} & 68.44 & 44.55 & 80.2 & \textbf{65.18} \\
    \midrule
    % 50 Clients
    \multirow{3}{*}{\rotatebox{90}{50 Clients}} & \textbf{Baseline (CART)} & 82.6 & 69.3 & 81.12 & 71.52 & 62.93 & 43.81 & 77.85 & 62.69 \\ 
    % & 77.972 & 68.698 & 82.553 & 64.04 & 80.59 & 78.30 & 79.79 & 62.57 \\
    \cmidrule{2-10}
    & \textbf{ICDTA4FL-ID3} & \textbf{87.42} & \textbf{74.04} & 81.8 & 70.41 & 67.6 & \textbf{53.55} & \textbf{81.31} & \textbf{64.93}  \\ 
    & \textbf{ICDTA4FL-CAR)} & 82.6 & 69.3 & \textbf{83.65} & \textbf{74.56} & \textbf{71.08} & 50.78 & 80.27 & 64.02 \\ 
    \bottomrule
    \end{tabularx}
    \caption{IID distribution results of the ICDTA4FL-CART model against the ICDTA4FL-ID3 model. The baseline refers to the mean of the local trees built by the clients when using CART as the base model.}
    % \caption{Results of ICDTA4FL process using CART as local tree-type and using IID data-distribution against \alberto{ICDTA4FL process} when using ID3 as local tree-type. The baseline refers to the mean of the local trees built by the clients when using CART as the base model.}
    \label{tab:cartcomparacioniid}
    \end{footnotesize}
\end{table}

\subsection{The ICDTA4FL-CART model: Results and Analysis}\label{subsec:resultsCART}

In this section, we present the results obtained by the proposed ICDTA4FL process when clients train a CART tree model locally. As shown in Section \ref{sec:background}, there is no method in the state-of-the-art that builds only a CART tree model between the clients, so we can not make a comparison. Furthermore, we do not compare the ICDTA4FL-CART model with other models that used bagging and boosting techniques, as those techniques improve the single-tree performance, and here we are building a single tree.\\

Instead, in this section we compare the ICDTA4FL-CART model, against the ICDTA4FL-ID3 model. To use the CART, it is necessary to preprocess the Nursery and the Car datasets, as they are categorical, so we have to transform their features into numerical features. Additionally, the Adult dataset is preprocessed to ensure all characteristics become numeric. \\

Tables \ref{tab:cartcomparacioniid} and \ref{tab:cartcomparacionnoiid} present the results of the ICDTA4FL process when CART is used as the base model versus when ID3 is used as the base model. Table \ref{tab:cartcomparacioniid} shows the results when there is an IID data distribution, and Table \ref{tab:cartcomparacionnoiid} shows the results when there is a non-IID data distribution.\\

\vspace{-.5cm}
\begin{table}[!bht]
    \centering
    \begin{footnotesize}
    \begin{tabularx}{\linewidth}{
    @{}r@{\hphantom{10}}X@{\hphantom{10}}r@{\hphantom{10}}r@{\hphantom{10}}r@{\hphantom{10}}r@{\hphantom{10}}r@{\hphantom{10}}r@{\hphantom{10}}r@{\hphantom{10}}r@{}
    }
    \toprule
    % Cabecera
    &  & \multicolumn{2}{c}{\textbf{Nursery}} & \multicolumn{2}{c}{\textbf{Adult}} & \multicolumn{2}{c}{\textbf{Car}} & \multicolumn{2}{c}{\textbf{Credit2}}\\ \midrule
    & & \textbf{Acc} & \textbf{F1} & \textbf{Acc} & \textbf{F1} & \textbf{Acc} & \textbf{F1} & \textbf{Acc} & \textbf{F1} \\ \midrule
    % 2 Clients
    \multirow{3}{*}{\rotatebox{90}{2 Clients}} & \textbf{Baseline (CART)} & 87.44 & 63.74 & 84.82 & 76.33 & 83.93 & 55.48 & \textbf{81.87} & \textbf{68.3} \\
    % & 81.904 & 73.491 & 90.87 & 69.452 & 87.322 & 86.323 & 81.096 & 66.809 \\ 
    \cmidrule{2-10}
    & \textbf{ICDTA4FL-ID3} & \textbf{91.45} & \textbf{71.45} & 81.66 & 73.36 & \textbf{86.25} & \textbf{60.93} & 80.96 & 67.55  \\
    & \textbf{ICDTA4FL-CART} & 87.65 & 63.09 & \textbf{85.06} & \textbf{76.78} & 86.13 & 58.8 & 81.5 & 64.97 \\ 
    \midrule
    % 5 Clients
    \multirow{3}{*}{\rotatebox{90}{5 Clients}} & \textbf{Baseline (CART)} & 87.07 & 65.52 & 84.49 & 76.27 & 85.06 & 56.26 & 81.66 & 67.2 \\
    \cmidrule{2-10}
    & \textbf{ICDTA4FL-ID3} & \textbf{91.75} & \textbf{75.31} & 82.45 & 73.81 & 79.95 & 45.76 & 80.4 & 64.7  \\
    & \textbf{ICDTA4FL-CART} & 88.72 & 66.81 & \textbf{85.22} & \textbf{77.53} & \textbf{87.55} & \textbf{62.89} & \textbf{82.08} & \textbf{68.05} \\
    \midrule
    % 10 Clients
    \multirow{3}{*}{\rotatebox{90}{10 Clients}} & \textbf{Baseline (CART)} & 86.30 & 64.9 & 84.14 & 75.56 & 82.25 & 54.71 & 80.7 & 66.17 \\
    \cmidrule{2-10}
    & \textbf{ICDTA4FL-ID3} & \textbf{91.52} & \textbf{75.08} & 80.39 & 68.75 & 79.74 & 54.02 & \textbf{84.28} & \textbf{71.53}  \\
    & \textbf{ICDTA4FL-CART} & 88.76 & 66.69 & \textbf{85.14} & \textbf{77.06} & \textbf{87.02} & \textbf{65.46} & 81.28 & 66.42 \\ 
    \midrule
    % 20 Clients
    \multirow{3}{*}{\rotatebox{90}{20 Clients}} & \textbf{Baseline (CART)} & 85.58 & 65.59 & 83.49 & \textbf{75.02} & 68.13 & 35.67 & 79.68 & 64.50 \\
    \cmidrule{2-10}
    & \textbf{ICDTA4FL-ID3} & \textbf{88.4} & \textbf{69.89} & 78.98 & 67.86 & \textbf{77.99} & \textbf{53.89} & 77.58 & 53.66  \\
    & \textbf{ICDTA4FL-CART} & 86.36 & 66.29 & \textbf{84.26} & 74.42 & 68.85 & 49.05 & \textbf{79.67} & \textbf{62.8} \\ 
    \midrule
    % 50 Clients
    \multirow{3}{*}{\rotatebox{90}{50 Clients}} & \textbf{Baseline (CART)} & 83.15 & 69.62 & 81.55 & 72.35 & 64.16 & 45.55 & 77.75 & 61.98 \\
    \cmidrule{2-10}
    & \textbf{ICDTA4FL-ID3} & \textbf{85.35} & \textbf{72.45} & 78.04 & 52.43 & \textbf{70.64} & 49.44 & \textbf{81.05} & \textbf{63.8}  \\
    & \textbf{ICDTA4FL-CART} & 76.7 & 61.72 & \textbf{81.94} & \textbf{72.78} & 69.6 & \textbf{50.63} & 78.37 & 61.04 \\ 
    \bottomrule
    \end{tabularx}
    \caption{Non-IID distribution results of the ICDTA4FL-CART model against the ICDTA4FL-ID3 model. The baseline refers to the mean of the local trees built by the clients when using CART as the base model.}
    % \caption{Results of ICDTA4FL process using CART as local tree-type and using a non-IID data-distribution against ICDTA4FL process when using ID3 as local tree-type. The baseline refers to the mean of the local trees built by the clients when using CART as the base model.}
    \label{tab:cartcomparacionnoiid}    
    \end{footnotesize}
\end{table}

Tables \ref{tab:cartcomparacioniid} and \ref{tab:cartcomparacionnoiid} contain the results of ICDTA4FL when using CART as the local tree versus the ICDTA4FL process when using ID3 as the local tree. These tables show that ICDTA4FL-ID3 works better in both numerical and categorical datasets when there is a higher number of clients training the model, but ICDTA4FL-CART performs better results with numerical datasets in scenarios with fewer clients.\\
% \newpage
\subsection{Analysis based on the number of clients}\label{subsec:ernc}
% \subsection{\alberto{Number of clients participating in training}}

A key aspect of FL is its ability to improve model performance through collaboration between different data owners. In this investigation, we examine the impact of diverse numbers of participating clients on the performance of the global model, and the results are shown in Tables \ref{tab:id3comparacioniid}, \ref{tab:id3comparacionnoiid}, \ref{tab:cartcomparacioniid} and \ref{tab:cartcomparacionnoiid}.\\

The number of participating clients goes from 2 to 50, with all the data split among the clients without sharing any instance. Thus, when the number of clients increases, each client has less data to train its local model, so, the number of trees that might add noise to the global model will be higher, affecting the performance of the global model.\\

The ICDTA4FL-ID3 model slightly decreases the performance of the global model when increasing the number of clients, but it keeps being higher than the state-of-the-art as shown in Tables \ref{tab:id3comparacioniid} and \ref{tab:id3comparacionnoiid}.\\

The ICDTA4FL-CART model can keep having a higher performance because of the filter selected, where we only keep DTs that might not introduce much noise to the global model, showing robustness when increasing the number of participant clients, as shown in Tables \ref{tab:cartcomparacioniid} and \ref{tab:cartcomparacionnoiid}.

\subsection{Robustness of percentile filtering along the ICDTA4FL process}\label{subsec:filter}

In Section \ref{sec:proposal} we expose that the ICDTA4FL process filters some DTs, that don't surpass a threshold, limiting the number of selected trees to build the global tree. We select by default the mean as a threshold, but other metrics can be used, such as percentile, median, mode, etc., being more restrictive with the threshold. Filtering is essential in the ICDTA4FL process, as it helps eliminate the number of trees that may add noise to the global DT when the number of clients increases, due to the low quality of extracted DTs.\\
\begin{figure}[H]
    \centering
    \includegraphics[scale=.8]{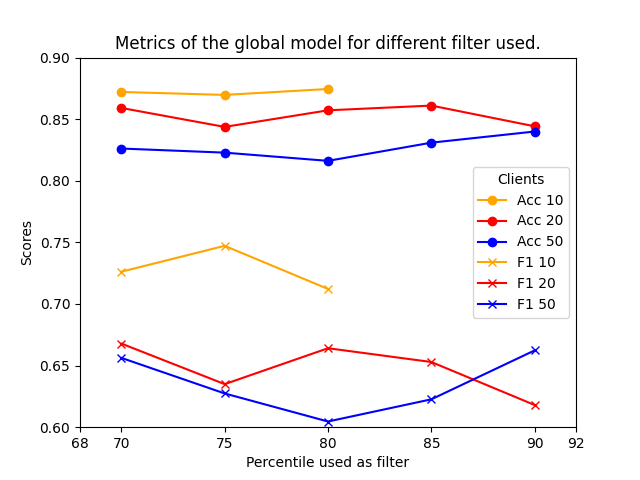}
    \caption{Exploring the impact of adjusting the filter on Accuracy and Macro-F1 scores in a Non-IID distribution on the Nursery Dataset using the ICDTA4FL-ID3 model. The percentile-based filter ignores the trees with metrics that do not surpass such percentile when constructing the global DT.}%Utilizing a percentile-based filter, only trees with metrics surpassing the percentile are utilized in constructing the global DT.}
    \label{fig:compfiltros}
\end{figure}
The reason for filtering the selected DTs is that, above all, in non-IID distributions there are clients with such low data, which trees will contribute with noise to the global model, harming its performance. To demonstrate this, we conducted an experiment using the ICDTA4FL process with the percentile as a filter. In Figure \ref{fig:compfiltros}, we show how the performance of the model varies in the Nursery dataset depending on the filter, based on different percentiles, being the results similar to the ones obtained when using the mean. Thus, the advantage of using a more restrictive filter is that it takes less time to build the model due to using fewer trees. \\

\subsection{Illustrative example on the behavior and interpretability}\label{subsec:interpretability}

DT methods are explainable by their architecture. They can be represented by association rules that manifest how input features influence predictions. This promotes trustworthiness by allowing stakeholders to verify the behavior of the model, understand the logic behind the decision, and check if it aligns with their expectations. We use this characteristic to discover reasons why the global DT performs better than the local DTs with ID3. % (see Tables \ref{tab:id3comparacioniid}-\ref{tab:id3comparacionnoiid}, $ICDTA4FL-ID3$ and Baseline respectively). 

\begin{table}[h!]
\begin{scriptsize}
\resizebox{\textwidth}{!} {
\begin{tabular}{|c|ccc|ll|}
\hline
\multicolumn{1}{|l|}{\multirow{2}{*}{Instance}} & \multicolumn{3}{c|}{Class}                                                 & \multicolumn{2}{c|}{Explanation}                                                                                                                                                                                                                                           \\ \cline{2-6} 
\multicolumn{1}{|l|}{}                          & \multicolumn{1}{c|}{Original} & \multicolumn{1}{c|}{$LocalDT_{i}$} & GlobalDT & \multicolumn{1}{c|}{$LocalDT_{i}$}                                                                                                            & \multicolumn{1}{c|}{GlobalDT}                                                                                                  \\ \hline
1                                               & \multicolumn{1}{c|}{1}        & \multicolumn{1}{c|}{2}          & 1        & \multicolumn{1}{l|}{\begin{tabular}[c]{@{}l@{}}\{x7==priority, \\   \hspace{0.2em} x1==improper, \\   \hspace{0.2em} x0==pretentious\}\end{tabular}}                     & \begin{tabular}[c]{@{}l@{}}\{x7==priority, \\   \hspace{0.2em} x1==improper, \\  \hspace{0.2em} x0==pretentious,\\   \hspace{0.2em} x5==convenient\}\end{tabular}          \\ \hline
2                                               & \multicolumn{1}{c|}{3}        & \multicolumn{1}{c|}{1}          & 3        & \multicolumn{1}{l|}{\begin{tabular}[c]{@{}l@{}}\{x7==recommended, \\   \hspace{0.2em} x1==very\_crit, \\   \hspace{0.2em} x6==slightly\_prob, \\   \hspace{0.2em} x3==2\}\end{tabular}} & \begin{tabular}[c]{@{}l@{}}\{x7==recommended, \\   \hspace{0.2em} x1==very\_crit, \\   \hspace{0.2em} x6==slightly\_prob, \\  \hspace{0.2em} x4==less\_conv\}\end{tabular} \\ \hline
3                                               & \multicolumn{1}{c|}{1}        & \multicolumn{1}{c|}{3}          & 1        & \multicolumn{1}{l|}{\begin{tabular}[c]{@{}l@{}}\{x7==recommended, \\  \hspace{0.2em} x1==critical, \\  \hspace{0.2em} x0==great\_pret, \\  \hspace{0.2em} x3==1\}\end{tabular}}      & \begin{tabular}[c]{@{}l@{}}\{x7==recommended, \\   \hspace{0.2em} x1==critical, \\   \hspace{0.2em} x0==great\_pret, \\  \hspace{0.2em} x3==1\}\end{tabular}               \\ \hline
\end{tabular}
}
\caption{Classification and explanation provided by the local DT of client $C_i$ ($LocalDT_{i}$) and the global DT ($GlobalDT$) models for three instances of the Nursery dataset using the ICDTA4FL-ID3 model.}
\label{tab:explainability}
    
\end{scriptsize}
\end{table}

We analyze the performance of the global model versus the local models by analyzing the class they predict and their associated rules. Table \ref{tab:explainability} presents three instances of the Nursery dataset that have been poorly classified by the local model and well classified by the global model. The first instance becomes well classified since the global model includes a condition on a characteristic ($x_5$) into the antecedent of the decision rule. The second instance becomes well classified since the global model replaces a condition of a characteristic ($x_3$) with a condition of another characteristic ($x_4$). The third instance becomes well classified by the global model, even having associated the same rule decision that the local model because it better defines the decision boundaries.

%% SECTION 6 - CONCLUSIONS AND FUTURE WORK
\section{Conclusions}\label{sec:conclusions}

In this paper, we propose a novel aggregation process for DTs in an FL environment namely ICDTA4FL. The ICDTA4FL process surpasses the state-of-the-art methods that build a single tree, as shown in Sections \ref{subsec:resultsID3} and \ref{subsec:resultsCART}. The ICDTA4FL process is tree-independent, capable of aggregating different types of DTs such as ID3 and CART while keeping the interpretability and the structure of the DTs used. ICDTA4FL has been evaluated in an HFL environment, in both IID and Non-IID distributions over 4 datasets, and varying the number of clients. We have used ID3 and CART as the base model, getting the ICDTA4FL-ID3 and the ICDTA4FL-CART models.\\ 

Therefore, we conclude that ICDTA4FL process presents the following advantages:
\begin{itemize}
    \item It can work with different types of DT, as shown in Section \ref{sec:exresanlysis}, being capable of adapting to the problem being solved.
    \item It achieves excellent performance in both IID and Non-IID distributions, improving the local models from the nodes.
    \item It keeps the hierarchical structure and interpretability for the original DTs used to build the global model. Thus, ICDTA4FL process is capable of generalizing the knowledge of the nodes, allowing us to understand the behavior behind the model.
\end{itemize}

In future work, we aim to extend the ICDTA4FL process to also work in other federated scenarios such as VFL, where there are multiple approaches for ensemble DT models \cite{federatedforest_vfl_2020, opboost2022acm}.
\section*{Acknowledgments}
This work is supported by the Recovery, Transformation and Resilience Plan, funded by the European Union (Next Generation Funds)
% This work is part of the grant ``Convenio de colaboración entre la Universidad de Granada y la S.M.E. Instituto Nacional de Ciberseguridad de España M.P., S.A." for the promotion of strategic cybersecurity projects in Spain, funded by the S.M.E. Instituto de Ciberseguridad de España M.P., S.A. and by the European Union - NextGenerationEU.

%\section*{Acknowledgements}

%Our work has been supported by the research project TIN2017-89517-P and by a research scholarship (FPU18/05989), given to the author Juan Luis Su\'arez by the Spanish Ministry of Science, Innovation and Universities.

%\section*{Compliance with ethical standards}

%\section*{Declaration of competing interest}

%The authors declare that there is no conflict of interest.

%\newpage
% Bibliography
%\section*{References} % En overleaf lo pone solo, en local es necesario (?)
% \biboptions{sort&compress}
\bibliography{main}

\end{document}